\documentclass{article}

\usepackage{microtype}
\usepackage{graphicx}
\usepackage{subfigure}
\usepackage{booktabs}

\usepackage{hyperref}

\usepackage[accepted]{icml2025}

\usepackage{amsmath}
\usepackage{amssymb}
\usepackage{mathtools}
\usepackage{amsthm}
\usepackage{balance}

\usepackage[capitalize,noabbrev]{cleveref}

\usepackage{wrapfig}

\theoremstyle{plain}

\theoremstyle{definition}

\theoremstyle{remark}

\usepackage[textsize=tiny]{todonotes}

\usepackage[utf8]{inputenc}
\usepackage[T1]{fontenc}    
\usepackage{url}            
\usepackage{amsfonts}       
\usepackage{nicefrac}       
\usepackage{xcolor}         
\usepackage{tabu}
\usepackage{multirow}
\usepackage{color}
\usepackage{amsmath,amssymb,amsfonts}
\usepackage{pifont}
\usepackage{subcaption}     
\usepackage{pgfplots}
\usepackage{enumitem}
\usepackage{bm}
\usepackage{diagbox}
\usepackage{float}
\usepackage{array}
\usepackage{xspace}
\usepackage{hyperref}

\newcommand{\methodname}{MiNT\xspace}
\pgfplotsset{compat=1.17}

\icmltitlerunning{MiNT: Multi-network Training for Transferrable Temporal Graph Models}

\begin{document}

\twocolumn[
\icmltitle{
MiNT: Multi-Network Training for Transfer Learning on Temporal Graphs
}

\icmlsetsymbol{equal}{*}

\begin{icmlauthorlist}
\icmlauthor{Kiarash Shamsi}{equal,uom}
\icmlauthor{Tran Gia Bao Ngo}{equal,uom}
\icmlauthor{Razieh Shirzadkhani}{equal,mila,mcgill}
\icmlauthor{Shenyang Huang}{mila,mcgill}
\icmlauthor{Farimah Poursafaei}{mila,mcgill}
\icmlauthor{Poupak Azad}{uom}
\icmlauthor{Reihaneh Rabbany}{mila,mcgill,cifar}
\icmlauthor{Baris Coskunuzer}{uot}
\icmlauthor{Guillaume Rabusseau}{mila,udm}
\icmlauthor{Cuneyt Gurcan Akcora}{ucf}

\end{icmlauthorlist}

\icmlaffiliation{uom}{Department of Computer Science, University of Manitoba}
\icmlaffiliation{mila}{Mila - Quebec AI Institute}
\icmlaffiliation{mcgill}{School of Computer Science, McGill University}
\icmlaffiliation{uot}{University of Texas at Dallas}
\icmlaffiliation{udm}{DIRO, Université de Montréal}
\icmlaffiliation{cifar}{CIFAR AI Chair}
\icmlaffiliation{ucf}{AI Initiative - University of Central Florida}


\icmlcorrespondingauthor{Kiarash Shamsi}{shamsik1@myumanitoba.ca}
\icmlcorrespondingauthor{Cuneyt Akcora}{cuneyt.akcora@ucf.edu}

\icmlkeywords{Machine Learning, ICML}

\vskip 0.3in
]

\printAffiliationsAndNotice{\icmlEqualContribution} 

\newcommand{\etc}{et al.\xspace}
\renewcommand{\vec}[1]{\ensuremath{\mathbf{#1}}}
\newcommand{\vecs}[1]{\ensuremath{\mathbf{\boldsymbol{#1}}}}
\newcommand{\mat}[1]{\ensuremath{\mathbf{#1}}}
\newcommand{\mats}[1]{\ensuremath{\mathbf{\boldsymbol{#1}}}}
\newcommand{\ten}[1]{\mat{\ensuremath{\boldsymbol{\mathcal{#1}}}}}

\newcommand{\ah}[1]{{{\textcolor{blue}{\textbf{AH: }}}{\textcolor{orange}{#1}}}}
\newcommand{\ks}[1]{{{\textcolor{blue}{\textbf{KS: }}}{\textcolor{blue}{#1}}}}
\newcommand{\bao}[1]{{{\textcolor{blue}{\textbf{Bao: }}}{\textcolor{blue}{#1}}}}
\newcommand{\rs}[1]{{{\textcolor{blue}{\textbf{RS: }}}{\textcolor{purple}{#1}}}}
\newcommand{\fp}[1]{{{\textcolor{blue}{\textbf{FP: }}}{\textbf{\textcolor{violet}{#1}}}}}
\newcommand{\gr}[1]{{{\textcolor{red}{\textbf{GR: }}}{\textcolor{red}{#1}}}}
\newcommand{\BC}[1]{{{\textcolor{blue}{\textbf{BC: }}}{\bf \textcolor{teal}{#1}}}}

\newcommand{\ca}[1]{{{\textcolor{blue}{\textbf{CA: }}}{\textcolor{violet}{#1}}}}
\newtheorem{definition}{Definition}
\newcommand{\name}{TGS\xspace}
\newcommand{\tempt}[1]{\textcolor{red}{#1}}
\newcommand{\note}[1]{\textcolor{red}{#1}}

\newcommand{\revised}[1]{\textcolor{blue}{#1}}

\newcommand{\training}{TGS-train\xspace}
\newcommand{\cmt}[1]{{\color{gray}{\texttt{//} #1}}}

\begin{abstract}

Temporal Graph Learning (TGL) has become a robust framework for discovering patterns in dynamic networks and predicting future interactions. While existing research has largely concentrated on learning from individual networks, this study explores the potential of learning from multiple temporal networks and its ability to transfer to unobserved networks. To achieve this, we introduce \textbf{Temporal Multi-network Training (\methodname)}, a novel pre-training approach that learns from multiple temporal networks. With a novel collection of 84 temporal transaction networks, we pre-train TGL models on up to 64 networks and assess their transferability to 20 unseen networks. 
Remarkably, MiNT achieves state-of-the-art results in zero-shot inference, surpassing models individually trained on each network. Our findings further demonstrate that increasing the number of pre-training networks significantly improves transfer performance. This work lays the groundwork for developing Temporal Graph Foundation Models, highlighting the significant potential of multi-network pre-training in TGL. Our code is available at \url{https://anonymous.4open.science/r/MiNT}
 
\end{abstract}

\section{Introduction}

Temporal graph learning has gained significant attention for its ability to model dynamic networks with evolving relationships, effectively addressing challenges that static graph representations cannot overcome \cite{longa2023graph}. Inspired by the remarkable success of pre-trained models in natural language processing (NLP)~\cite{bubeck2023sparks, brown2020language, rasul2024lagllama} and computer vision (CV)~\cite{radford2021learning, awais2023foundational}, there has been a growing interest to develop Graph Foundational Models (GFMs)~\cite{mao2024position}. These models employ a \textit{pre-train and transfer} strategy, enabling Graph Neural Networks (GNNs) to leverage large-scale datasets for pre-training and efficiently transfer to new tasks with minimal supervision~\cite{wang2024llms,beaini2023towards}. However, in temporal graph learning, existing literature focuses on training and testing on a single temporal network~\cite{huang2023tgb,rossi2020TGN,Aldo2020EvolveGCN}, therefore learning the evolution of only one network. Inspired by the success of large pre-trained models, we explore two fundamental questions about temporal graph models:
(1)~Can temporal graph models benefit from learning across multiple networks? 
(2)~Can temporal graph models effectively transfer the learned knowledge from pre-training datasets to unseen networks within the same domain?

\begin{figure*}[h!]
    \centering
    \includegraphics[width=0.85\textwidth]{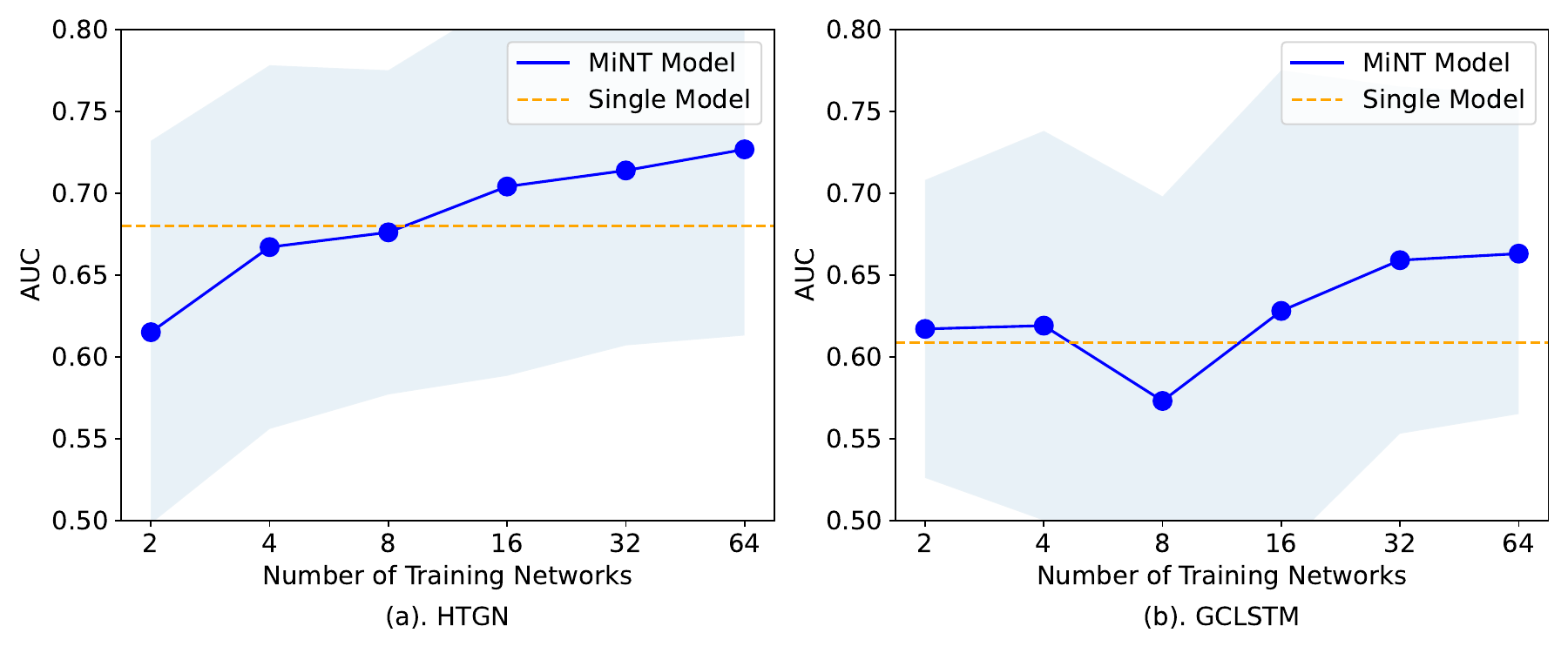}
    \caption{\textbf{Scaling behavior of MiNT on unseen networks.} Zero-shot inference performance of MiNT~(\emph{Multi-Network Model}) on unseen networks,  compared with standard training of individual networks~(\emph{Single Model}). The base TGNN models are (a) HTGN, and (b) GC-LSTM. A single model is trained and tested on each test network, while MiNT performs zero-shot inference on each test network. The metric is the average ROC AUC over 20 test networks.} 
   
    \label{fig:fig_1}
    
    \vspace{-0.4cm} 
\end{figure*}

To investigate these questions, we selected Ethereum-based cryptocurrency networks as a testbed and constructed a novel collection of 84 temporal transaction networks, comprising 3 million nodes and 19 million edges. This dataset facilitates the study of transferability and multi-network training, providing a robust foundation for future research in temporal graph learning. Next, we propose the first algorithm to pre-train temporal graph neural networks (TGNNs) across multiple networks, developing a novel approach named \textbf{Temporal Multi-network Training (\methodname)}. 
By training \methodname on up to 64 cryptocurrency networks, we assess its transfer learning performance compared to state-of-the-art TGNN models.
Figure~\ref{fig:fig_1} shows how our multi-network models, compared to two SOTA models, namely HTGN and GCLSTM, perform as we train them for graph property prediction tasks. 
The average performance of \methodname on twenty unseen token networks improves as more training data is provided, outperforming single models trained on individual networks. Our results demonstrate that \methodname consistently performs on par with or even better individually trained models, underscoring its transformative potential to drive progress in transfer learning for temporal graph learning. 
Our contributions are:

$\bullet$ \textbf{Multi-network Training Algorithm:} We introduce \textit{\methodname-train}, the first algorithm to train TGNNs across multiple networks simultaneously, expanding possibilities for temporal graph learning. 

$\bullet$ \textbf{Positive Dataset Scaling:} Our \methodname model demonstrates significant improvement as it learns from an increasing number of networks (up to 64), highlighting the advantages of multi-network learning on temporal graphs.

$\bullet$ \textbf{Superior Transferability:}  \methodname achieves competitive zero-shot inference on 20 unseen token networks, matching or even outperforming SOTA on $14$ out of $20$ unseen networks. This demonstrates exceptional generalization and transfer capabilities.

$\bullet$ \textbf{Extensive Temporal Network Dataset:} We present a dataset of 84 labeled temporal graphs derived from token transaction networks, designed to support future research on scaling, transferability, and multi-network learning. 

\textbf{Reproducibility.} Our code is available on \href{https://anonymous.4open.science/r/MiNT}{https://anonymous.4open.science/r/MiNT}. The \methodname datasets are available on \href{https://www.dropbox.com/sh/hsjrzu4x0d2x4a0/AAAjAqkd8kO4RkjFMPmNdo1ma?dl=0}{Dropbox}.

\begin{figure*}[t!]
    \centering
   
    \includegraphics[width=1\linewidth]{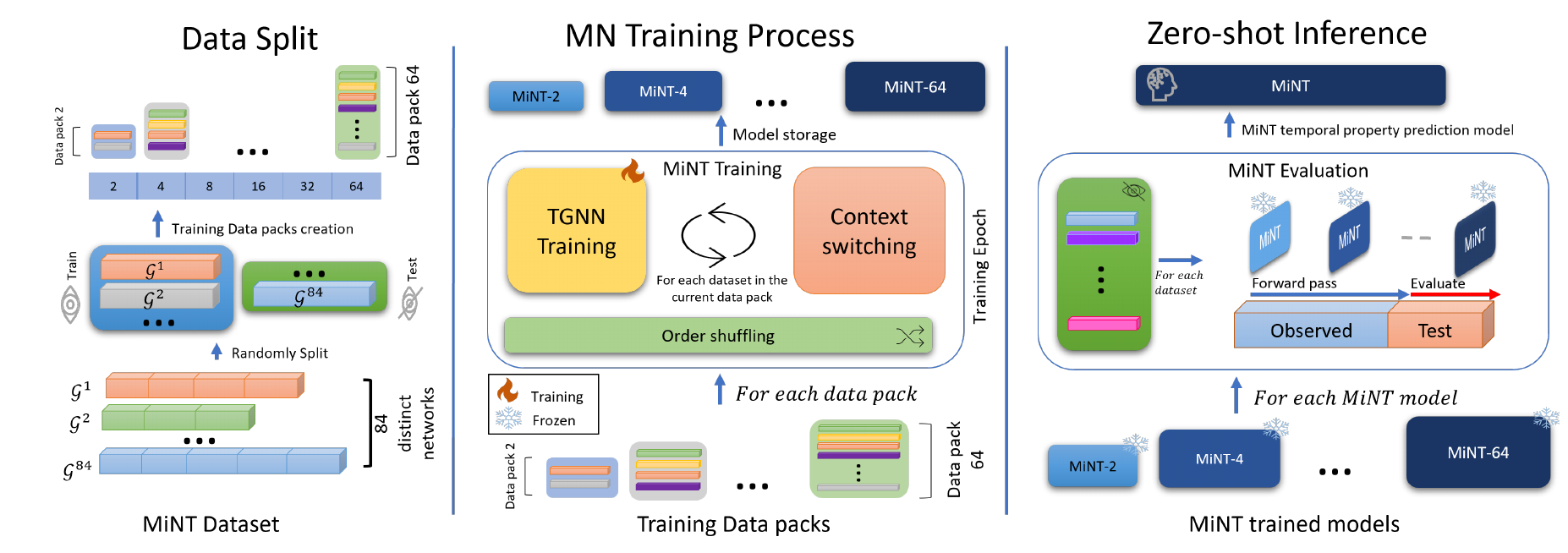}
    \caption{\textbf{\methodname framework}. Temporal graphs are preprocessed to generate discrete-time snapshots. Next, the multi-network training pipeline leverages these snapshots to train TGNNs across multiple networks for zero-shot inference on unseen temporal networks. 
    } 
    \label{fig:system_overview_MINT}
    
    \vspace{0cm} 
\end{figure*}

\section{Related Work}\label{sec:relate}

\paragraph{Temporal Graph Learning.}
Temporal graph neural networks (TGNNs) have shown promising performance in tasks such as link prediction and node classification. Current literature~\cite{rossi2020TGN,menglin2021HTGN,chen2022GCLSTM} focuses on learning from a single network and partitioning the network into a training set and a test set chronologically. Thus, the objective is to extract patterns from the observed temporal graph and then predict its evolution in the future. 
In the inductive settings, TGNNs predict accurately for novel nodes from the \textit{same} network that were not observed in the training set~\cite{da2020TGAT,wang2021inductive}. In this work, we propose the first algorithm to effectively learn from multiple temporal graphs and focus on the temporal graph property prediction task. Additionally, we test the transferability of TGNNs on novel networks, unobserved during training. 

\citet{kazemi2020representation} categorized temporal graphs into Discrete Time Dynamic Graphs (DTDGs) and Continuous Time Dynamic Graphs (CTDGs). In this work, we focus on DTDGs, since the temporal graph property prediction task is more appropriately defined over graph snapshots~\cite{shamsi2024graphpulse}. 
DTDG methods process each graph snapshot sequentially and use recurrent modules to learn temporal dependencies. For example, GCLSTM~\cite{chen2022GCLSTM} stacks a graph CNN for feature extraction and an LSTM cell for temporal reasoning. In addition, leveraging the power of hyperbolic geometry, HTGN maps the temporal graph into hyperbolic space and utilizes hyperbolic graph neural networks and hyperbolic-gated recurrent neural networks to model the evolving dynamics.
Another SOTA for temporal graph property prediction is GraphPulse~\cite{shamsi2024graphpulse}, which leverages topological data analysis tools to extract information from each snapshot. However, all DTDG models are designed to train and test on a single network, while in this work, we explore multi-network training.

\paragraph{Graph Foundation Models.} Recently, \citet{xia2024anygraph} outlined the challenges associated with building a graph foundation model as structural heterogeneity, feature heterogeneity, fast adaptation and achieving neural scaling law. Neural scaling law is often used to categorize the relationship between model performance and factors in model training and design such as the number of parameters, the size of the training set and the amount of computation required~\cite{rosenfeld2020constructivePrediction,kaplan2020scalingLawLModel,abnar2022limitsPreTrain}. \citet{liu2024graphScalingLaws} investigated neural scaling laws for static graphs by observing the performance of GNNs given the increase in the model size and training set size. Graph foundation models such as GraphAny~\cite{zhao2024graphany} and AnyGraph~\cite{xia2024anygraph} focus on designing architectures that allow for easy adaption to unseen networks. There are also approaches to build domain-specific foundation models such as those for molecular graphs~\cite{beaini2023towards,sypetkowski2024scalability,klaser2024minimol}. In this work, we provide a novel collection of temporal graphs, necessary for building future TG foundation models while demonstrating the significant benefit of pre-training TGNNs on multiple networks for transferability to unseen networks.

\textbf{Zero-shot Inference.}
Zero-shot learning has emerged as a powerful approach to enable pre-trained models to extrapolate prediction on unseen data from datasets used in the pre-training stage. 
Inspired by recent advancements in zero-shot learning and the power of pre-trained models in LLM~\cite{rasul2024lagllama} and CV~\cite{vladan2024label},
\citet{wang2024llms} introduced TEA-GLM, a novel framework that aligns GNN representations with LLM token embeddings by a linear projector. The incorporation of LLMs and GNNs enables zero-shot inference on unseen graphs.
Additionally, \citet{xia2024opengraph} proposed OpenGraph, an initiative promoting transparency, reproducibility, and community-driven advancements in graph representation learning. In this work, our proposed \methodname algorithm allows pre-trained TGNNs to achieve zero-shot inference without the need for fine-tuning or modifications, relying directly on the inference pass of a frozen pre-trained model.

\section{Background} \label{sec:background}

This section provides the background for temporal graph learning and the temporal graph property prediction task.

In this work, we focus on DTDGs which represent the temporal evolution of the graph in discrete graph snapshots. 
\begin{definition}[\textbf{Discrete Time Dynamic Graphs}] 
DTDGs represent a network as a sequence of graph snapshots denoted as \(\mathcal{G} = \{\mathcal{G}_{1}, \mathcal{G}_{2}, \mathcal{G}_{3}, \ldots, \mathcal{G}_{n}\}\). 
Each snapshot \(\mathcal{G}_{t} = (\mathcal{V}_{t}, \mathcal{E}_{t}, \mat{X}_{t}, \mat{Y}_{t})\) corresponds to the graph at timestamp \(t\), where \(\mathcal{V}_{t}\) and \(\mathcal{E}_{t}\) are the sets of nodes and edges, respectively. \(\mat{X}_{t}\) and \(\mat{Y}_{t}\) are the node and edge feature matrices, respectively. A collection of DTDGs is defined as \(D = \{ \mathcal{G}^1, \mathcal{G}^2,\ldots, \mathcal{G}^m \}\), where \(m\) is the number of DTDGs.
\end{definition}

\textbf{Temporal Graph Property Prediction.} This task focuses on forecasting a specific property of a temporal graph over a future time interval in a DTDG. Formally, given a DTDG \(\mathcal{G}\), we define a target time interval \([t_{k + \delta_1}, t_{k + \delta_2}]\), where \(\delta_1\) and \(\delta_2\) are non-negative integers and \(\delta_1 \leq \delta_2\). At a specific time \(t_k\), the objective is to predict the values of the chosen graph property within the specified future interval \([t_{k + \delta_1}, t_{k + \delta_2}]\). This task is particularly valuable for analyzing the scalability and evolution of temporal graphs. For instance, in financial domains, predicting the growth or shrinkage of the network can provide insights into market dynamics and investor behavior. Moreover, leveraging the DTDGs framework for graph property prediction can uncover critical patterns in domains like cryptocurrency token networks~\cite{cuneyt2019bitcoinHeist}.

\section{\methodname: Temporal Multi-network Training}
\label{sec:methodology}

In this section, we introduce our Temporal Multi-network Training~(\methodname) algorithm, an innovative multi-network pre-training framework designed to be applied to any TGNN architecture for DTDG. By leveraging \methodname pre-training, the base TGNN model can now transfer to unseen networks for zero-shot inference. Figure~\ref{fig:system_overview_MINT} provides an overview of our \methodname framework, illustrating the process from dataset curation to the model pre-training stage, and finally to zero-shot inference on test networks.

 \begin{algorithm}[tb]
 \begin{algorithmic}[1]
\caption{Multi-Network Training for Temporal Graphs} 
\label{alg:foundational}
\STATE {\bfseries Input:} A Temporal Graph Dataset $D = \{ \mathcal{G}^1, \mathcal{G}^2,\ldots, \mathcal{G}^m \}$, where $\mathcal{G}^i = \{\mathcal{G}_{1}^i, \mathcal{G}_{2}^i, \ldots, \mathcal{G}_{n_i}^i\}$, $\emph{m} = $ Number of networks in training,  a $\mathbf{TGNN}$ model, and a graph property prediction $\mathbf{Decoder}$
\FOR {each $epoch$}
\STATE // \textit{Order shuffling}
\STATE Shuffle ($D$) 
\FOR{each network $\mathcal{G}^i \in D$}
\STATE // \textit{Context switching}
\STATE Initialize historical embeddings $\mathcal{H}_0$ (reset) 
\FOR{$t=1,2,\cdots,n_i$}
\STATE // \textit{Train on snapshot $\mathcal{G}^i_{t}$}
\STATE $\mathcal{H}_{t}=\mathbf{TGNN}(\mathcal{G}^i_{t},\mathcal{H}_{t-1} )$ 
\STATE $\hat{y}_{t} = \mathbf{Decoder}(\mathcal{H}_{t})$
\STATE $\mathcal{L} = \mathbf{Loss}(y_{t},\hat{y}_{t})$\;
\STATE Backpropagation\;
\ENDFOR
\STATE Evaluate on the validation snapshots of $\mathcal{G}^i$ \;
\ENDFOR

\STATE Average validation results across all datasets to select the best model\;
\STATE Save the best model for inference \;

\ENDFOR

\end{algorithmic}
\end{algorithm}

\subsection{Multi-network Training}

Existing temporal graph learning models typically train on a single temporal graph, limiting their ability to capture similar behaviors and generalize across different networks~\cite{rossi2020TGN,menglin2021HTGN}. 
In contrast, we consider a classical learning scenario where a training dataset of $m$ temporal graphs $D = \{ \mathcal{G}^1, \mathcal{G}^2,\ldots, \mathcal{G}^m \}$ is drawn identically and independently~(IID) from an unknown distribution, and the learned model is evaluated on a test set of unseen temporal networks drawn from the same distribution.

Our \methodname algorithm trains across multiple temporal graphs by modifying a SOTA single network training model with two crucial steps: \textit{shuffling} and \textit{context switching}. As explained below, these steps render the algorithm network-agnostic, capable of learning from various temporal graphs to generalize effectively to unseen networks. 
Algorithm~\ref{alg:foundational} shows \methodname-train approach in detail. As the first step, we load the list of temporal graphs $D$, where each temporal graph $\mathcal{G}^i$ is represented as a sequence of snapshot $\{\mathcal{G}^i_{1}, \mathcal{G}^i_{2}, \ldots, \mathcal{G}^i_{n_i}\}$. For each epoch,  we shuffle the orders of the list of datasets $D$ to preserve the IID assumption of neural network training.

\textbf{Order Shuffling.} Since previous works focus on training models on a single network for temporal tasks, we incorporate a shuffling step at each epoch to facilitate training on multiple networks and enable inference on unseen networks. The randomized ordering of networks during training at each epoch is important because it helps prevent the model from learning spurious correlations that could arise if the data were presented in a fixed order. Shuffling the datasets promotes randomness in the training process, contributing to more robust and generalizable model performance.
Sequentially, for each dataset $\mathcal{G}^i$, we first initialize the historical embeddings, then train the model end-to-end 
on each dataset $\mathcal{G}^i$ in a similar manner of training a single model, and evaluate the performance on the corresponding validation set of dataset $\mathcal{G}^i$. After training on $m$ datasets from $D$, we compute the average validation results across these datasets. This average is used to select the best model, which is then used for inference. Early stopping is applied if needed. The importance of order shuffling is demonstrated in the ablation study presented in Table~\ref{tab:ablation-results}.
 
\textbf{Context Switching.} Many TGNNs store and utilize node embeddings or latent states from previous timestamps at later timestamps; we refer to those embeddings as \emph{historical embeddings} \cite{menglin2021HTGN,chen2022GCLSTM,Aldo2020EvolveGCN}. In Algorithm~\ref{alg:foundational}, this is represented  in line 10 
as 
$$ \mathcal{H}_{t}=\mathbf{TGNN}(\mathcal{G}^i_{t},\mathcal{H}_{t-1} ),$$
indicating that at time steps $t$, the temporal graph model takes as input both the current snapshot $\mathcal{G}^i_{t}$ and the \textit{latent state} $\mathcal{H}_{t-1}$ from the previous time step~(similar to RNNs).
Resetting historical embeddings at the beginning of each epoch~(line~7 
of Algorithm~\ref{alg:foundational}) is a key step in training a temporal model across multiple networks for several reasons. First, it helps prevent the model from carrying over biases or assumptions from one network to another, ensuring that it can adapt effectively to the unique characteristics of each network. Starting with fresh historical embeddings at the beginning of each epoch enables the model to learn the most relevant and up-to-date information from the current network, improving performance and generalization across different networks. This is equivalent to resetting the initial vector of recurrent neural networks at the beginning of each sequence.

\textbf{Zero-shot Inference.}
To evaluate the transferability of each multi-network model, we test the model on the test sets that are unseen by the models during the training phase. We first divide our networks into two disjoint sets, where one set is used for training obtained by randomly selecting 64 token networks, and the remaining 20 token networks are used to evaluate the performance. In the inference phase, we begin by loading all the weights of multi-network models, including the pre-trained encoder and decoder parameters, while initializing fresh historical embeddings. Then, we perform a single forward pass over the train and validation split to adapt the historical embeddings specific to the testing dataset.

\subsection{\methodname Datasets}

We introduce a diverse collection of large token networks derived from the real-world Ethereum blockchain~\cite{wood2014ethereum}, capturing authentic transaction patterns from 2017 to 2023. The dataset includes 84 distinct ERC20 token networks, each reflecting unique investor and transaction behaviors.  Importantly, these token networks operate independently, with varying start dates, durations, and evolution, making them valuable for analyzing shared and unique characteristics within temporal graph structures. The data extraction process and its structure are provided in Appendix~\ref{sec:dataset}.
We split the 84 token networks into two sets: 64 networks are used for training, while the remaining 20 networks, which are never seen during training, are reserved exclusively for zero-shot inference. This split ensures that the model's ability to generalize to entirely new networks can be effectively evaluated. \methodname \  contains a diverse set of dynamic graphs in terms of nodes, edges, and timestamps, which are shown in Figure~\ref{fig:data-stats}. Details on statistics are given in Appendix~\ref{appendix:data_stat}.
\begin{figure}[t]
    \centering
    \includegraphics[width=\linewidth]{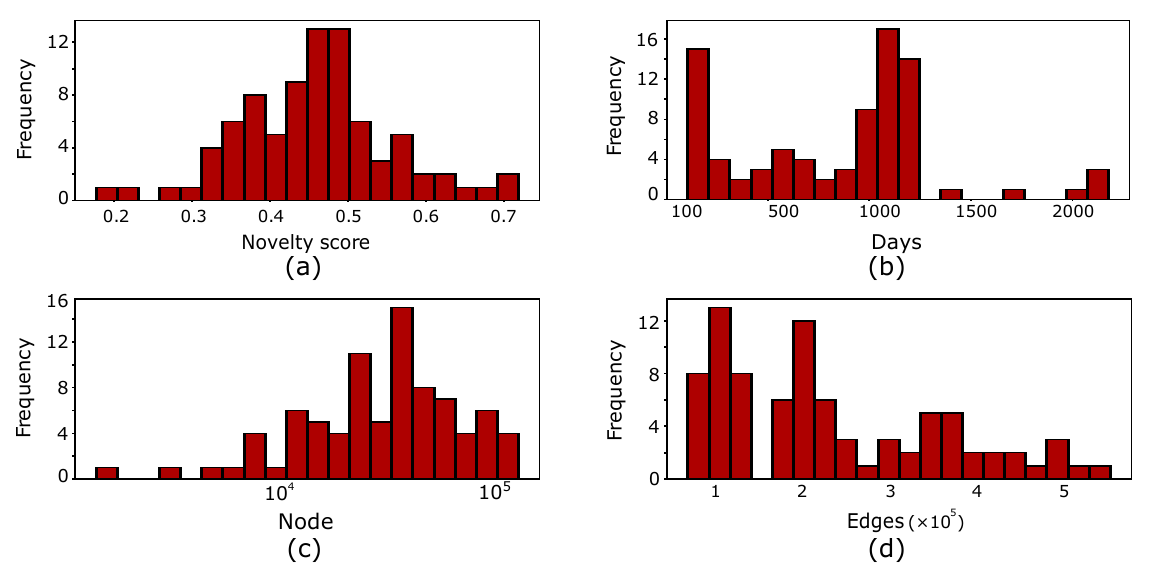}
    \caption{\textbf{Network statistics of \methodname \  networks}. (a) Novelty score, (b) number of days, (c) number of nodes, and (d) number of edges. }
    \label{fig:data-stats}
\vspace{-.3in}
\end{figure}

\section{Experiments} \label{sec:exps} 

In this section, we conduct empirical experiments to evaluate the transferability of our proposed \methodname approach to unseen test networks. Our experiments focus on the temporal graph property prediction task for predicting network growth or shrinkage~(see Section~\ref{sec:background}). As weekly forecasts are common in the financial context for facilitating financial decisions~\cite{kim2021predicting}, we set  $\delta_{1}=3$ and $\delta_{2} = 10$ for the temporal graph property prediction task, thus predicting the temporal graph property over weekly snapshots. 
In addition, to show that \methodname is agnostic to the underlying TGNN architecture, we select two widely-used TGNN models as base architecture, namely HTGN~\cite{menglin2021HTGN} and GCLSTM~\cite{chen2022GCLSTM}. Experimental results suggest that HTGN has superior performance overall; thus, we adapt this model for in-depth analysis and ablation studies.~\footnote{The default \methodname models refer to HTGN with our \methodname pre-training, for example in Figure~\ref{fig:htgn-log2-bar}} We also provide detailed experimental results and findings in Appendix~\ref{app:detailed_results}.

\begin{table*}[t]
    \centering

     \caption{\textbf{ROC AUC} scores of \methodname models~(ours, zero-shot inference on unseen test networks), single models~(trained on test networks), and persistence forecasts on test sets across three seeds. Best performance in \textbf{bold}, second best \underline{underlined}.} \label{tab:auc-result}

  \resizebox{.8\linewidth}{!}{%
      \begin{tabular}{l | c  c  c c c | c c}
    \toprule 

    &   \multicolumn{5}{c|}{\textbf{Standard Training on Individual Networks}}  &  \multicolumn{2}{c}{\textbf{\methodname-64 (Transfer)}} \\
       \textbf{Network} & PF & HTGN & GC-LSTM  & EvolveGCN & GraphPulse & GC-LSTM & HTGN \\ 
              \midrule
        WOJAK & 0.378 & 0.479\scriptsize $\pm$0.005 & 0.484\scriptsize $\pm$0.000 & 0.505\scriptsize $\pm$0.023
        & 0.467\scriptsize $\pm$0.030 & 
        \textbf{0.534\scriptsize $\pm$0.020} & \underline{0.524\scriptsize $\pm$0.027}  \\

        DOGE2.0 & 0.250 & \textbf{0.590\scriptsize $\pm$0.059} & 0.538\scriptsize $\pm$0.000 & \underline{0.551\scriptsize $\pm$0.022}
        & 0.384\scriptsize $\pm$0.180 & 
        \underline{0.551\scriptsize $\pm$0.022} & 0.538\scriptsize $\pm$0.038  \\

        EVERMOON & 0.241 & 0.512\scriptsize $\pm$0.023 & \textbf{0.562\scriptsize $\pm$0.179} & 0.451\scriptsize $\pm$0.046
        &\underline{0.519\scriptsize $\pm$0.130} & 
        0.494\scriptsize $\pm$0.047 & 0.517\scriptsize $\pm$0.039  \\

        QOM & 0.334 & 0.633\scriptsize $\pm$0.017 & 0.612\scriptsize $\pm$0.001 & 0.618\scriptsize $\pm$0.002
        & \textbf{0.775\scriptsize $\pm$0.011} & 
        0.618\scriptsize $\pm$0.004 & \underline{0.647\scriptsize $\pm$0.019}  \\

        SDEX & 0.423 & \textbf{0.762\scriptsize $\pm$0.034} & 0.720\scriptsize $\pm$0.002 & \underline{0.733\scriptsize $\pm$0.028}
        & 0.436\scriptsize $\pm$0.030 & 
        0.723\scriptsize $\pm$0.002 & 0.614\scriptsize $\pm$0.020  \\

        ETH2x-FLI & 0.355 & 0.610\scriptsize $\pm$0.059 & 0.670\scriptsize $\pm$0.009 & 0.688\scriptsize $\pm$0.010
        & {0.666\scriptsize $\pm$0.047} & 
        \underline{0.697\scriptsize $\pm$0.009} & \textbf{0.729\scriptsize $\pm$0.015}  \\

        BEPRO & 0.393 & 0.655\scriptsize $\pm$0.038 & 0.632\scriptsize $\pm$0.019 & 0.610\scriptsize $\pm$0.012
        & \textbf{0.783\scriptsize $\pm$0.003} & 
         {0.746\scriptsize $\pm$0.015} & \underline{0.782\scriptsize $\pm$0.003}   \\

        XCN & 0.592 & 0.668\scriptsize $\pm$0.099 & 0.306\scriptsize $\pm$0.092 & 0.512\scriptsize $\pm$0.067 
        & \underline{0.821\scriptsize $\pm$0.004} & 
        0.733\scriptsize $\pm$0.003 & \textbf{0.851\scriptsize $\pm$0.043}  \\

        BAG & 0.792 & 0.673\scriptsize $\pm$0.227 & 0.196\scriptsize $\pm$0.179 & 0.329\scriptsize $\pm$0.040
        & \textbf{0.934\scriptsize $\pm$0.020} & 
        {0.529\scriptsize $\pm$0.023} & \underline{0.931\scriptsize $\pm$0.028}  \\

        TRAC & 0.400 & 0.712\scriptsize $\pm$0.071 & 0.748\scriptsize $\pm$0.000 & 0.748\scriptsize $\pm$0.000 
        & \underline{0.767\scriptsize $\pm$0.001} & 
        0.742\scriptsize $\pm$0.004 & \textbf{0.785\scriptsize $\pm$0.008}   \\

        DERC & 0.353 & 0.683\scriptsize $\pm$0.013 & 0.703\scriptsize $\pm$0.022 & 0.669\scriptsize $\pm$0.009 
        & \underline{0.769\scriptsize $\pm$0.040} & 
        {0.696\scriptsize $\pm$0.011} & \textbf{0.798\scriptsize $\pm$0.027}  \\

        Metis & 0.423 & 0.715\scriptsize $\pm$0.122 & 0.646\scriptsize $\pm$0.023 & 0.688\scriptsize $\pm$0.027
        & \textbf{0.812\scriptsize $\pm$0.011} & 
        {0.697\scriptsize $\pm$0.013} & \underline{0.760\scriptsize $\pm$0.025}  \\

        REPv2 & 0.321 & 0.760\scriptsize $\pm$0.012 & 0.725\scriptsize $\pm$0.014 & 0.709\scriptsize $\pm$0.002
        & \textbf{0.830\scriptsize $\pm$0.001} & 
        {0.733\scriptsize $\pm$0.019} & \underline{0.789\scriptsize $\pm$0.020}  \\

        DINO & 0.431 & 0.730\scriptsize $\pm$0.195 & \textbf{0.874\scriptsize $\pm$0.028} & \underline{0.868\scriptsize $\pm$0.029}
        & {0.801\scriptsize $\pm$0.020} & 
         0.659\scriptsize $\pm$0.039 & 0.779\scriptsize $\pm$0.113 \\

        HOICHI & 0.374 & 0.807\scriptsize $\pm$0.047 & \textbf{0.857\scriptsize $\pm$0.000} & \underline{0.856\scriptsize $\pm$0.001}
        & 0.714\scriptsize $\pm$0.010 & 
        0.847\scriptsize $\pm$0.005 & 0.765\scriptsize $\pm$0.018   \\

        MUTE & 0.536 & 0.649\scriptsize $\pm$0.015 & 0.593\scriptsize $\pm$0.030 & 0.617\scriptsize $\pm$0.010
        & \textbf{0.779\scriptsize $\pm$0.004} & 
        0.636\scriptsize $\pm$0.003 & \underline{0.673\scriptsize $\pm$0.013}  \\

        GLM & 0.427 & \underline{0.830\scriptsize $\pm$0.029} & 0.451\scriptsize $\pm$0.003 & 0.501\scriptsize $\pm$0.033
        & 0.769\scriptsize $\pm$0.018 & 
        0.501\scriptsize $\pm$0.027 & \textbf{0.831\scriptsize $\pm$0.024}  \\

        MIR & 0.327 & 0.750\scriptsize $\pm$0.005 & 0.768\scriptsize $\pm$0.026 & 0.745\scriptsize $\pm$0.015
        & {0.689\scriptsize $\pm$0.097} & 
        \underline{0.788\scriptsize $\pm$0.022} & \textbf{0.836\scriptsize $\pm$0.016}  \\

        stkAAVE & 0.426 & {0.702\scriptsize $\pm$0.042} & 0.368\scriptsize $\pm$0.011 & 0.397\scriptsize $\pm$0.022
        & \textbf{0.743\scriptsize $\pm$0.006}& 
        0.650\scriptsize $\pm$0.028 & \underline{0.709\scriptsize $\pm$0.022}  \\

        ADX & 0.362 & \underline{0.769\scriptsize $\pm$0.018} & 0.723\scriptsize $\pm$0.002 & 0.718\scriptsize $\pm$0.004
        & \textbf{0.784\scriptsize $\pm$0.002} & 
        0.673\scriptsize $\pm$0.022 & 0.679\scriptsize $\pm$0.024   \\

        \midrule
        Top rank $\uparrow$  & 0 & 2 & 3  & 0  & \textbf{8} & 1 & \underline{6} \\ 
        \midrule
        Avg. rank $\downarrow$  & 6.60 & 3.60 & 4.30  & 4.30  & \underline{2.85} & 3.70 & \textbf{2.40} \\ 
       
\bottomrule
    \end{tabular} }
\end{table*}

\subsection{Contenders And Baselines}

For comparison, we include heuristics, single-network models, and our pre-trained \methodname models. We refer to models that are trained on only a single network (such as in existing literature) as \emph{single models}. 
For each temporal graph, we adopt a $70\%-15\%-15\%$ split ratio for the train, validation, and test sets, respectively, and during each epoch, the snapshots from a network are processed sequentially and chronologically by the model. 

We train every single model for a minimum of $100$ and a maximum of $250$ epochs with a learning rate set to $1.5 \times 10^{-3}$. We apply early stopping based on the AUC results on the validation set, with patience and tolerance set to $20$ and $5 \times 10^{-2}$, respectively.
We use Binary Cross-Entropy Loss for performance measurement and Adam \cite{kingma2015Adam} as the optimization algorithm. It is important to note that the graph pooling layer, performance measurement, and optimization algorithm are also shared by the multi-network model training setup.

\paragraph{Persistence Forecast.}
For our baseline model, we employ a naive approach similar to deterministic heuristics techniques, persistence forecast~\cite{salcedo2022}, for label prediction. We further provide details about the persistence forecast baseline in  \Cref{appendix:singlemodels}. Hyperparameter details are described in Appendix~\ref{ap:hyperparameters}.

\paragraph{Single Models.} We use four models from literature, including HTGN~\cite{menglin2021HTGN}, GCLSTM~\cite{chen2022GCLSTM}, EvolveGCN~\cite{Aldo2020EvolveGCN} and GraphPulse~\cite{shamsi2024graphpulse} as our baseline models trained from scratch on individual networks. We further explain each model in \Cref{appendix:singlemodels}.  
To address graph-level tasks, we add an extra graph pooling layer as the final layer. This layer, implemented as a Multi-Layer Perceptron (MLP), takes the mean of all node embeddings, concatenating with four snapshot features at the graph level (i.e., the mean of in-degree, the weight of in-degree, out-degree, and weight of out-degree) and outputs binary classification prediction. 

\paragraph{\methodname Models.}
We train six \textcolor{black}{multi-network} models, each with a different number of networks corresponding to $2^k$ datasets, where $k \in [1, 6]$. We name each \textcolor{black}{multi-network} model based on the number of datasets used in training; for example, \textcolor{black}{\methodname}-16 is trained with 16 datasets. The default TGNN architecture is HTGN, which shows superior performance, while GCLSTM architecture is also trained and discussed in Table~\ref{tab:auc-result}.

\paragraph{Computional Resource.} We ran all experiments on NVIDIA Quadro RTX 8000 (48G memory) with 4 standard CPU nodes (either Milan Zen 3  2.8 GHz and 768GB of memory each or Rome Zen 2, 2.5GHz and 256GB of memory each). Figure~\ref{fig:train-time} shows the training time for \methodname models which scales linearly to number of training networks.

 \subsection{Experimental Results}

\begin{figure*}[t]
    \centering
    
    \includegraphics[width=\textwidth]{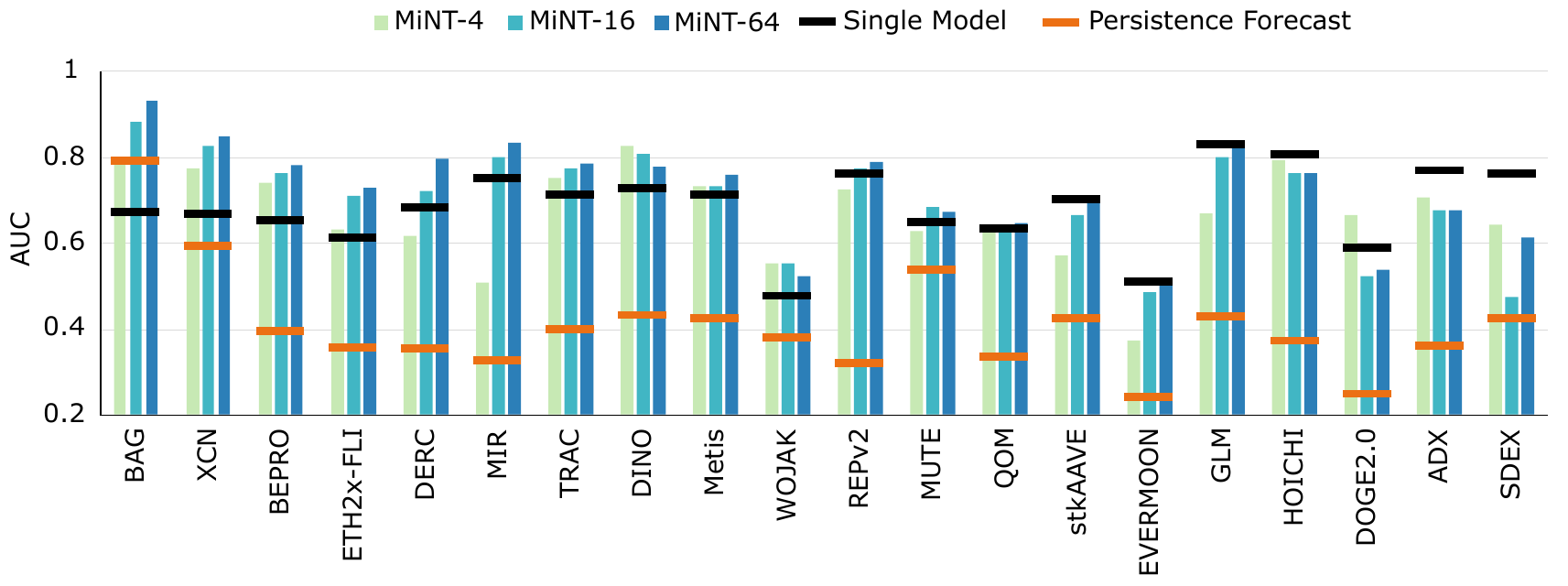}
    \caption{\textbf{MiNT Performance with Varying Training Scales.} Test AUC of MiNT models trained on 4, 16 and 64 networks and evaluated on unseen test datasets. We compare the performance with persistence forecast, and HTGN models trained and tested on each dataset.
    } \label{fig:htgn-log2-bar}
\end{figure*}

\paragraph{The Benefit Of \methodname Training.}

We present the performance of our \textcolor{black}{multi-network} models trained with datasets of varying sizes \textcolor{black}{and zero-shot inference tested} on \(20\) unseen test datasets. The results are reported with an average and standard deviation of three runs with different random seeds. We compare our results with five baseline single models: Persistence Forecast, GCLSTM, EvolveGCN, HTGN, and GraphPulse, which are trained and tested on the same \(20\) datasets. For visual clarity, Figure~\ref{fig:htgn-log2-bar} shows the AUC on test data results for \textcolor{black}{\methodname-4, \methodname-16 and \methodname-64} only as well as persistence forecasting and HTGN single model. We show the performance of all six \textcolor{black}{multi-network} models in Appendix Figure~\ref{fig:htgn-all}. Overall, an upward trend is observed in most datasets from \textcolor{black}{multi-network models} \(2\) to \(64\), such as in \textit{BAG}, \textit{MIR} and \textit{BEPRO} datasets, highlighting the power of larger multi-network models in temporal graph learning. In Figure~\ref{fig:htgn-log2-bar}, the \methodname-64 yields the best AUC in 16 out of $20$ test datasets. This result is significant because the multi-network models outperform the single models individually trained on these datasets. We detail the prediction performance of the models in Table \ref{tab:auc-result}, where we present the AUC values for both single-trained baselines and multi-network models, specifically \methodname-32 and \methodname-64, across various datasets. We also report the top rank, average rank, and win ratio for each model. The top rank indicates the number of datasets where a method ranks first. To calculate the average rank, we assign an AUC-based rank (ranging from $1$ to $8$) to every model across the 20 test datasets and compute the average. The win ratio represents the proportion of datasets where a model outperforms a single model.

Overall, \methodname-64 exhibits the best overall performance, achieving the state-of-the-art AUC performance in $6$ networks and top two performance in $14$ out of $20$ total test networks with zero-shot inference. 
Moreover, Appendix Table \ref{tab:train_set} indicates that the \methodname-64 also achieves superior performance on the test set of $38$ out of $64$ training networks when compared to the performance of single models. This demonstrates the strong generalizability and transferability of our \methodname-64 model.  While GraphPulse achieves the highest top rank of $8$, it relies on trained inference, unlike our multi-network models, which are based on zero-shot inference. Notably, training GraphPulse on each dataset is computationally expensive, while inference testing of our pre-trained \methodname-64 on all datasets takes only a few minutes. This makes the performance of \methodname-64 even more remarkable. Furthermore, despite trained models like HTGN or GCLSTM performing well on certain datasets, our \methodname-64 model consistently achieves competitive rankings across all datasets.

\begin{table}[h!]
\centering
\caption{Rank-based prediction performances.
 }
  \label{tab:rank-result}
  \resizebox{0.95\linewidth}{!}{%
    \begin{tabular}{l  c  c  c }
    \toprule 
    \toprule 
      Model  & Top rank $\uparrow$  & Avg. rank $\downarrow$  & Win ratio $\uparrow$\\ \midrule
      Persist. forecast & 0 & 7.9 & 0.00\\
      Single model& 3 &  4.35 & -\\
      \methodname-2& 0 & 6.15& 0.25\\
      \methodname-4& 2 & 4.35& 0.45\\
      \methodname-8& 1 & 4.45& 0.45\\
      \methodname-16& 1 & 3.45& 0.65\\
      \methodname-32& 2 & 3.20& 0.70\\
      \methodname-64& \textbf{11} & \textbf{2.15} & \textbf{0.80}\\
    \bottomrule
    \bottomrule
    \end{tabular}%
    }
   
\end{table}

\paragraph{Effect Of Scaling.}
In Table \ref{tab:rank-result}, we further compare the models by reporting the top rank, average rank, and win ratio for different configurations of the multi-network models. We observe a notable improvement in performance as the number of training networks increases. For instance, the average rank (lower is better) improves from $6.15$ for \methodname-2 to $2.15$ for \methodname-64, which signifies a roughly $50\%$ performance enhancement when scaling from two networks to sixty-four. The improvement in the win ratio is also substantial, with \methodname-64 achieving the highest win ratio of $0.80$, outperforming the other models in most datasets. This indicates that increasing the number of networks in a multi-network model significantly enhances its robustness and predictive power, particularly when compared to single models and smaller multi-network configurations. Overall, the Multi-network HTGN-based models have shown superior zero-shot performance and transferability ability. 

\begin{table*}[]
    \centering
    \caption{Ablation study results (AUC) demonstrating the impact of various training strategies on model performance.}
  \label{tab:ablation-results}

  \resizebox{0.8\linewidth}{!}{%
    \begin{tabular}{l  c  c  c  c  c}
    \toprule 
    \toprule 
      Model  & \methodname-4 $\uparrow$  & \methodname-8 $\uparrow$ & \methodname-16 $\uparrow$ & \methodname-32 $\uparrow$ & \methodname-64 $\uparrow$\\ \midrule
      Base Model & 0.667\scriptsize $\pm$0.111 & 0.676\scriptsize $\pm$ 0.099 & 0.704\scriptsize $\pm$ 0.115 & 0.714\scriptsize $\pm$ 0.107 & 0.727\scriptsize $\pm$ 0.114\\ 
      w/o Order shuffling & 0.647\scriptsize $\pm$0.113 & 0.643\scriptsize $\pm$0.117 & 0.690\scriptsize $\pm$ 0.105 & 0.708\scriptsize $\pm$ 0.099 & 0.694\scriptsize $\pm$ 0.109\\
      w/o Context Switching & 0.667\scriptsize $\pm$0.120 & 0.608\scriptsize $\pm$0.102 & 0.693\scriptsize $\pm$ 0.099 & 0.677\scriptsize $\pm$ 0.098 & 0.688\scriptsize $\pm$ 0.095\\
    \bottomrule
    \bottomrule
    \end{tabular}%
    }
    \vskip -0.2in
\end{table*}

\paragraph{\methodname Ablation Study.}
We conducted an ablation study for the \methodname-train algorithm to assess the effects of \emph{context switching} and \emph{order shuffling}. Models are trained the same as multi-network model training setup and tested on the 20 unseen test datasets. The average results are presented in Table \ref{tab:ablation-results}. Training different multi-network models without resetting memory revealed that persistent memory across epochs negatively impacts generalization, emphasizing the importance of reset mechanisms to reduce overfitting. Additionally, we explored the necessity of shuffling data by fixing the order of training networks. The observed performance decline indicated that incorporating randomness to \methodname is vital for improving the model's robustness and generalizability.

\paragraph{Effect Of Data Selection.}  We investigate the effect of data selection on the performance of \methodname models trained with different training data packs. As the first work on multi-network training for temporal graphs, we explore the importance of our dataset selection process. To avoid any bias, we randomly sampled the training datasets from the 64 available networks. We conducted an  empirical experiment to examine the impact of dataset selection on training \methodname models. In this experiment, we choose three disjoint sets of datasets (data pack A, B, and C) for training \methodname-2, \methodname-4, \methodname-8, and \methodname-16 and two disjoint sets of datasets (data pack A, B) for training \methodname-32. Using disjoint data packs ensures that each model is trained on unique data, eliminating any overlap that could obscure the results. We then test our models on 20 unseen test datasets.

\begin{figure}[]
    \centering
    \includegraphics[width=.95\linewidth]{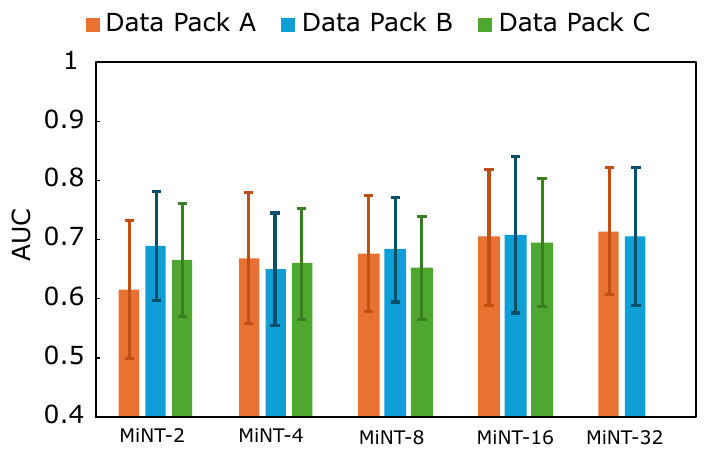}
    \caption{Effect of Data Selection on model performance.}
    \label{fig:data_packs}
     \vspace{-.2in}
\end{figure}

\balance

As shown in Figures~\ref{fig:data_packs}, as the number of training networks increases, the \textcolor{black}{multi-network} model performance increases while the variance between different choices of training networks decreases. However, the difference between models that use the same number of datasets diminishes as we move from models of 2 to 32 datasets. We observe that smaller models (i.e., \methodname-2) have a higher variance when compared to larger models (i.e., \methodname-64); in addition, the model performance also increases from small to large models. For example, \methodname-64 outperforms \methodname-32 on 16 out of 20 datasets. 

\begin{figure}
\vspace{.5cm}
    \centering
    \includegraphics[width=.95\linewidth]{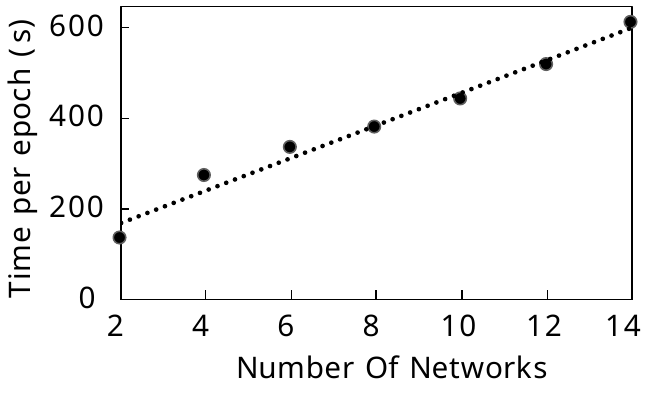}
    \caption{Time per epoch for training multi-network models.}
    \label{fig:train-time}
    \vspace{-1pt}

\end{figure}

 \textbf{Time Complexity Analysis.} The \methodname-train algorithm has the same complexity as training the single model across all the training networks. Specifically, for the best performing HTGN-based model, the time complexity using \methodname-train is $O(m \cdot (N_{max} dd' + d'|\mathcal{E}_{max}|))$ where $m$ is the number of training networks, $N_{max}$ is set to the maximum number of nodes of networks in the training set, $d$ and $d'$ are the dimensions of the input and output features while $|\mathcal{E}_{max}|$ is the maximum number of edges in a snapshot. Empirically, we observe that the \methodname-training time scales linearly to the number of networks as seen in Figure~\ref{fig:train-time} where we report the time per epoch for each multi-network model.

\section{Conclusion}

In this work, we seek to address the question: given a collection of observed temporal graphs, can we predict the evolution of an unseen network within the same domain? We find that it is indeed possible to learn from temporal networks in the same domain and predict future trends for unseen networks.
First, we collected and released a collection of \(84\) temporal networks for the temporal graph property prediction task. These datasets serve as the foundation for studying neural scaling laws and foundation models on temporal graphs. Next, to learn from a large number of temporal graphs, we present~\methodname-train, the first algorithm for training TGNNs across multiple temporal networks.
Experimentally, we show that the neural scaling law also applies to temporal graphs; in particular, the more training networks are used, the better the model performance on unseen test networks. In addition, our trained multi-network models can outperform single models trained on individual test networks. 
Our empirical observations show the high potential of training foundational models on temporal graphs.
We believe our \methodname will pave the way for advancements in temporal graph foundation models, providing valuable resources that the community can utilize.

\section*{Broader Impacts}
The proposed research on pre-training temporal graph neural networks across multiple networks has the potential to advance the field of machine learning and its applications significantly. By introducing methodologies to enhance the scalability and transferability of TGNNs, this work could revolutionize areas like network security, financial fraud detection, and real-time social network analysis, where dynamic and adaptive models are essential. The publicly available dataset of 84 Ethereum-based temporal networks will serve as a valuable resource for the research community, fostering innovation and collaboration. Furthermore, the principles of multi-network pre-training introduced here can inspire analogous advances in other temporal data domains, such as healthcare, transportation, and climate science. This research opens up a new direction in training generalizable temporal graph models that, for the first time, can be trained on distinct temporal networks, paving the way for Temporal Graph Foundation Models. 

This work also introduces a set of Ethereum transaction token networks, which are publicly available to users who have the  necessary resources, such as fast SSDs, large RAM, and ample disk space, to synchronize Ethereum clients and manually extract blocks. Additionally, all Ethereum data is accessible on numerous Ethereum explorer sites such as \href{https://etherscan.io/}{etherscan.io}. An Ethereum user's privacy depends on whether personally identifiable information (PII) is associated with any of their blockchain address, which serves as account handles and are considered pseudonymous. If such PII were obtained from other sources, our datasets could potentially be used to link Ethereum addresses. However, real-life identities can only be discovered using IP tracking information, which we neither have nor share. Our data does not contain any PII. Furthermore, we have developed a request to exclude an address from the dataset.

\bibliography{references}
\bibliographystyle{icml2025}

\clearpage
\appendix

\centerline{\bf \Large Appendix}
\section{MiNT Datasets} \label{sec:dataset}

Numerous graph benchmark datasets have been introduced to advance research within the temporal graph learning community. \citet{Poursafaei2022BetterEvaluation} introduced six dynamic graph datasets while proposing visualization techniques and novel negative edge sampling strategies to facilitate link prediction tasks of dynamic graphs. Following the good practice from OGB~\citep{weihua2020OGB}, \cite{huang2023tgb} introduced TGB, which provides automated and reproducible results with a novel standardized evaluation pipeline for both link and node property prediction tasks. However, these datasets belong to different domains, making them unsuitable for studying the scaling laws of neural network models trained with a large number of datasets from the same domain. 
\cite{li2024evaluating} provide a temporal benchmark for evaluating graph neural networks in link prediction tasks, though their focus does not extend to training on multiple networks. Conversely, the Live Graph Lab dataset by \cite{zhang2023livegraph} offers a temporal dataset and benchmark, employed for tasks like temporal node classification using TGNNs. This work aims to explore multi-network training and understand the transferability across temporal graphs. Therefore, we curate a collection of temporal graphs rather than focusing on individual ones as in prior work.

\subsection{Datasets Extraction}
We utilize a dataset of temporal graphs sourced from the Ethereum blockchain~\cite{wood2014ethereum}. In this section, we will describe Ethereum, explain our data pipeline, and conclude by defining the characteristics of the resulting dataset.

\textbf{Ethereum and ERC20 Token Networks.}  \label{sec:raw_data}
 We create our transaction network data by first installing an Ethereum node and accessing the P2P network by using the Ethereum client  Geth (\url{https://github.com/ethereum/go-ethereum}). Then, we use Etherum-ETL (\url{https://github.com/blockchain-etl/ethereum-etl}) to parse all ERC20 tokens and extract asset transactions. We extracted more than sixty thousand ERC20 tokens from the entire history of the Ethereum blockchain. However, during the lifespans of most token networks, there are interim periods without any transactions. Additionally, a significant number of tokens live for only a short time span. To avoid training data quality challenges, we use 84 token networks with at least one transaction every day during their lifespan and are large enough to be used as a benchmark dataset for \textcolor{black}{multi-network} model training.

 \textbf{Temporal Networks.} Each token network represents a distinct temporal graph, reflecting the time-stamped nature of its transactions. In these networks, nodes (addresses), edges (transactions), and edge weights (transaction values) evolve over time, capturing the dynamic behavior of the network. Additionally, these networks differ in their start dates and durations, introducing further variation in their evolution. While each token network operates independently with its own set of investors, they exhibit common patterns and behaviors characteristic of transaction networks. These similarities allow the model to learn and generalize from these patterns across different networks. Collecting temporal graphs from different ERC20 token networks allows for comparative analysis, uncovering in-common patterns and unique behaviors. This strengthens the model's ability to generalize and improves its robustness.

\begin{figure}[b]
 \centering
      \includegraphics[width=\linewidth]{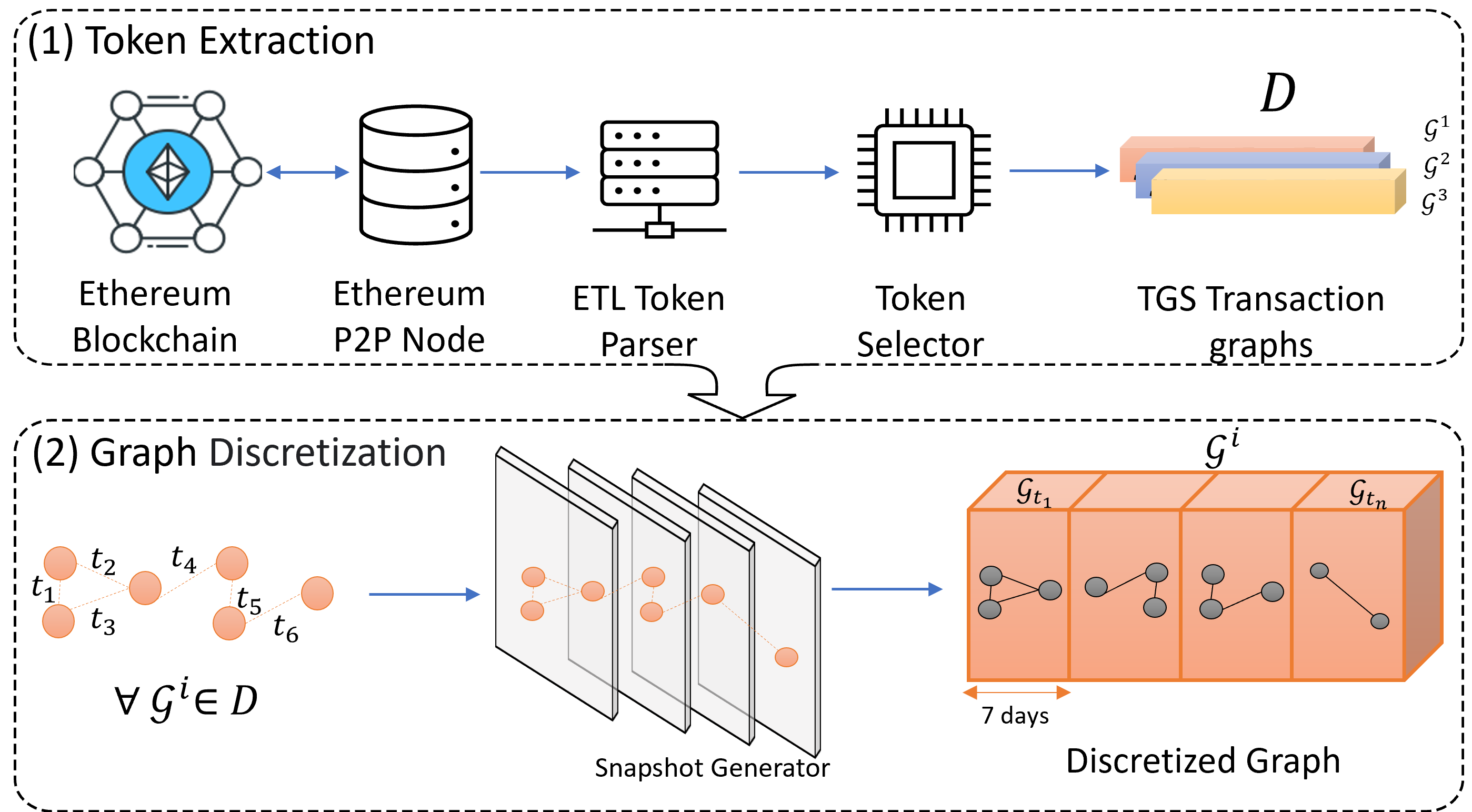}
      \caption{\textbf{\methodname \ data processing overview.} (1) \underline{\textit{Token extraction}}: extracting the token transaction network from the Ethereum node. (2) \underline{\textit{Discretization}}: creating weekly snapshots to form discrete time dynamic graphs.}
     \label{fig:system_overview}
 \end{figure}

Figure \ref{fig:system_overview} illustrates the \methodname \  overview from dataset extraction and discretizing graph networks for model training step.

\begin{table*}[htbp]
  \centering
  \caption{All token networks' statistics.}
  \label{tab:TGS_all_stats}
  \resizebox{\linewidth}{!}{%
    \begin{tabular}{lrrclll | lrrclll}
    \toprule 
  
      Token  & \#Node & \#Transaction & \#Snapshots (days) & Growth rate & Novelty & Surprise & Token &\#Node & \#Transaction & \#Snapshots (days) & Growth rate & Novelty & Surprise  \\ \midrule

      ARC & 11325 & 70968 & 606 & 0.43 & 0.32 & 0.88 & Metis & 52586 & 343141 & 907 & 0.44 & 0.48 & 0.89\\
CELR & 65350 & 235807 & 1691 & 0.49 & 0.56 & 0.96 & cDAI & 52753 & 358050 & 1437 & 0.45 & 0.46 & 0.9\\
CMT & 86895 & 205961 & 309 & 0.45 & 0.72 & 0.92 & BITCOIN & 34051 & 347054 & 178 & 0.48 & 0.39 & 0.63\\
DRGN & 113453 & 341849 & 2164 & 0.44 & 0.57 & 0.97 & INJ & 60472 & 312822 & 1113 & 0.46 & 0.52 & 0.98\\
GHST & 35156 & 180955 & 1146 & 0.43 & 0.51 & 0.93 & MIM & 23038 & 269366 & 885 & 0.44 & 0.4 & 0.89\\
INU & 8556 & 66315 & 154 & 0.27 & 0.41 & 0.59 & GLM & 53385 & 234912 & 1080 & 0.5 & 0.53 & 0.96\\
IOTX & 63079 & 288469 & 1993 & 0.45 & 0.56 & 0.99 & Mog & 14590 & 240680 & 107 & 0.37 & 0.38 & 0.55\\
QSP & 117977 & 299671 & 2178 & 0.45 & 0.67 & 0.99 &DPI & 40627 & 234246 & 1150 & 0.49 & 0.5 & 0.86\\

REP & 83282 & 224843 & 346 & 0.46 & 0.69 & 0.96 &LINA & 45342 & 227147 & 1144 & 0.45 & 0.46 & 0.95\\

RFD & 23208 & 173695 & 169 & 0.3 & 0.39 & 0.6 & Yf-DAI & 22466 & 226875 & 1158 & 0.42 & 0.31 & 0.87\\

TNT & 88247 & 316352 & 1216 & 0.43 & 0.55 & 0.93 & BOB & 42806 & 212099 & 199 & 0.35 & 0.48 & 0.73\\

TRAC & 71667 & 299181 & 2110 & 0.46 & 0.54 & 0.97 & RGT & 35277 & 211932 & 1110 & 0.44 & 0.46 & 0.98\\

RLB & 28033 & 240291 & 129 & 0.43 & 0.49 & 0.76 & TVK & 42539 & 208082 & 1062 & 0.41 & 0.48 & 0.93\\
 
steCRV & 19079 & 211538 & 1033 & 0.45 & 0.53 & 0.9 & RSR & 50645 & 205906 & 659 & 0.47 & 0.62 & 0.91\\

ALBT & 63042 & 434881 & 1152 & 0.43 & 0.44 & 0.89 & WOJAK & 34341 & 198653 & 201 & 0.37 & 0.48 & 0.73\\

POLS & 128159 & 554705 & 1132 & 0.45 & 0.61 & 0.94 & ANT & 36517 & 200262 & 1107 & 0.47 & 0.46 & 0.93\\

SWAP & 69230 & 509769 & 1213 & 0.46 & 0.45 & 0.79 & LADYS & 37486 & 192176 & 181 & 0.37 & 0.52 & 0.79\\

SUPER & 83299 & 502030 & 986 & 0.47 & 0.46 & 0.85 & ETH2x-FLI & 11008 & 199088 & 965 & 0.47 & 0.28 & 0.84\\

RARI & 87186 & 502960 & 1207 & 0.43 & 0.47 & 0.91 & TURBO & 38638 & 189048 & 189 & 0.33 & 0.48 & 0.72\\

KP3R & 39323 & 493258 & 1102 & 0.43 & 0.33 & 0.88 & REPv2 & 39061 & 191367 & 1194 & 0.48 & 0.5 & 0.97\\

MIR & 79984 & 444998 & 1066 & 0.45 & 0.43 & 0.92 & NOIA & 29798 & 185528 & 1133 & 0.46 & 0.37 & 0.7\\

aUSDC & 23742 & 475680 & 1067 & 0.46 & 0.4 & 0.73 & 0x0 & 21531 & 182430 & 283 & 0.51 & 0.46 & 0.81\\

LUSD & 25852 & 430473 & 943 & 0.48 & 0.36 & 0.87 & PSYOP & 25450 & 168896 & 169 & 0.32 & 0.39 & 0.59\\

PICKLE & 28498 & 430262 & 1149 & 0.48 & 0.34 & 0.69 & ShibDoge & 40023 & 134697 & 680 & 0.43 & 0.53 & 0.8\\

DODO & 47046 & 390443 & 1131 & 0.47 & 0.45 & 0.91 & ADX & 14567 & 123755 & 1188 & 0.44 & 0.4 & 0.91\\

YFII & 43964 & 391984 & 1196 & 0.44 & 0.44 & 0.96 & BAG & 11860 & 122634 & 298 & 0.31 & 0.44 & 0.87\\

STARL & 71590 & 369913 & 856 & 0.46 & 0.48 & 0.86 & QOM & 21757 & 118292 & 598 & 0.46 & 0.41 & 0.81\\
 
LQTY & 34687 & 374230 & 943 & 0.45 & 0.34 & 0.91 & BEPRO & 26521 & 120261 & 1132 & 0.46 & 0.48 & 0.87\\

FEG & 118294 & 367584 & 1007 & 0.4 & 0.62 & 0.92 & AIOZ & 29231 & 119926 & 947 & 0.43 & 0.49 & 0.89\\

AUDIO & 91218 & 362685 & 1108 & 0.45 & 0.58 & 0.95 & PRE & 40476 & 118625 & 1113 & 0.5 & 0.55 & 0.86\\

OHM & 45728 & 377068 & 690 & 0.43 & 0.46 & 0.88 & CRU & 19990 & 117712 & 1144 & 0.5 & 0.43 & 0.95\\

WOOL & 16874 & 351178 & 716 & 0.41 & 0.18 & 0.41 & POOH & 27245 & 111641 & 193 & 0.26 & 0.49 & 0.69\\

DERC & 24277 & 111205 & 824 & 0.45 & 0.49 & 0.83 & aDAI & 13648 & 187050 & 1068 & 0.45 & 0.46 & 0.82\\

stkAAVE & 37355 & 110924 & 1128 & 0.42 & 0.57 & 0.71 & ORN & 44010 & 239451 & 1134 & 0.46 & 0.47 & 0.87\\

BTRFLY & 8450 & 108371 & 453 & 0.48 & 0.34 & 0.44 & DOGE2.0 & 7664 & 79047 & 123 & 0.45 & 0.38 & 0.66\\

SDEX & 9127 & 104869 & 240 & 0.41 & 0.44 & 0.75 & HOICHI & 5075 & 77361 & 436 & 0.36 & 0.32 & 0.71\\

XCN & 20085 & 104185 & 607 & 0.46 & 0.42 & 0.84 & EVERMOON & 7552 & 79868 & 163 & 0.24 & 0.35 & 0.52\\

HOP & 37004 & 102650 & 514 & 0.41 & 0.6 & 0.88& MUTE & 12426 & 82345 & 977 & 0.43 & 0.46 & 0.95\\

MAHA & 18401 & 96180 & 749 & 0.43 & 0.47 & 0.91 & crvUSD & 2950 & 88647 & 174 & 0.61 & 0.37 & 0.73\\

DINO & 15837 & 94140 & 358 & 0.44 & 0.44 & 0.74& SLP & 6675 & 95368 & 1151 & 0.43 & 0.36 & 0.91\\

bendWETH & 1454 & 96898 & 593 & 0.51 & 0.21 & 0.51& sILV2 & 12838 & 92905 & 611 & 0.4 & 0.34 & 0.48\\

PUSH & 14501 & 93103 & 936 & 0.46 & 0.38 & 0.83 & SPONGE & 25852 & 90468 & 184 & 0.31 & 0.66 & 0.81\\

\bottomrule
    \end{tabular}%
    }
\end{table*}

\subsection{Dataset Statistics } \label{appendix:data_stat}

Our \methodname \  dataset is a collection of \(84\) ERC20 token networks derived from Ethereum from 2017 to 2023. Each token network is represented as a dynamic graph, in which each address and transaction between addresses are a node and directed edge, respectively. The biggest  \methodname \  token network contains \(128,159\) unique addresses and \(554,705\) transactions, while the smallest token network has \(1,454\) nodes.

\begin{figure*}[htbp]
    \centering
    \subfigure[\scriptsize Unique Nodes of training set]{
        \includegraphics[width=0.25\linewidth]{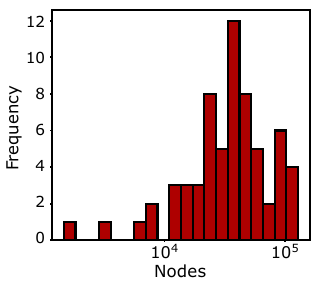}
        \label{fig:node_distribution_training}
        }\quad
    \subfigure[\scriptsize Transaction of training set]{
         \includegraphics[width=0.25\linewidth]{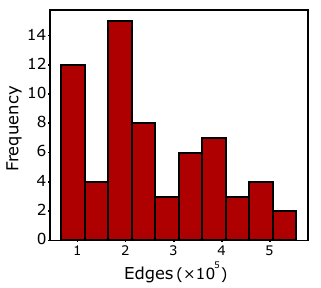}
        \label{fig:edges_distribution_training}
    }\quad
    \subfigure[\scriptsize Length in days of training set]{
        \includegraphics[width=0.25\linewidth]{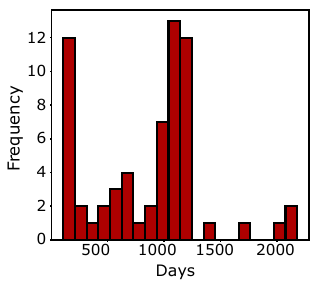}
        \label{fig:length_distribution_training}
        }
        
    \subfigure[\scriptsize Unique Nodes of testing set]{
        \includegraphics[width=0.25\linewidth]{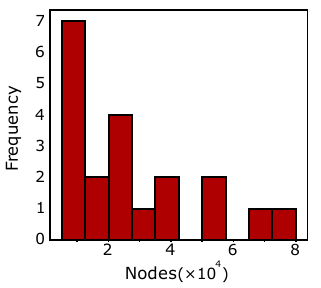}
        \label{fig:node_distribution_testing}
    }\quad
    \subfigure[\scriptsize Transactions of testing set]{
         \includegraphics[width=0.25\linewidth]{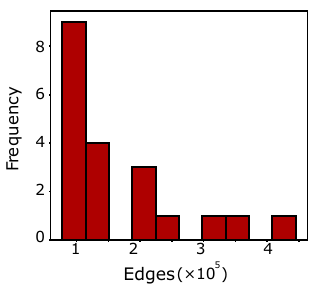}
        \label{fig:edges_distribution_testing}
    }\quad
    \subfigure[\scriptsize Length in days of testing set]{
        \includegraphics[width=0.25\linewidth]{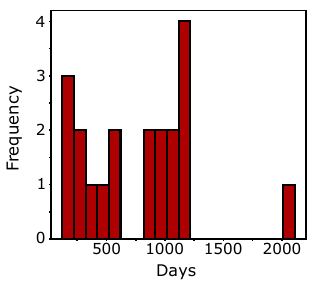}
        \label{fig:length_distribution_testing}
    }    
    \caption{Distribution of the characteristics of the datasets over training and testing sets.}
    \label{fig:TGS_training_testing_characteristic}
\end{figure*}

Figure~\ref{fig:data-stats} shows that most networks have more than 10k nodes and over 100k edges. The lifespan of  \methodname \  networks varies from \(107\) days to \(6\) years, and there exists at least one transaction each day. Figure~\ref{fig:data-stats}.a shows the novelty scores, i.e., the average ratio of unseen edges in each timestamp, introduced by \cite{Poursafaei2022BetterEvaluation}. Figure~\ref{fig:data-stats} shows that most of the $84$ networks have novelty scores greater than $0.3$, indicating that each day sees a considerable proportion of new edges in these token networks. We adopt a \(70-15-15\) split of train-test-validation for each token network and calculate the surprise score~\cite{Poursafaei2022BetterEvaluation}, which indicates the number of edges that appear only in the test data. As Table~\ref{tab:TGS_all_stats} shows, the token networks have quite high surprise values with an average of $0.82$.  We also provide the node, edge and length distribution for train and test sets separately in Figure~\ref{fig:TGS_training_testing_characteristic}. Overall, train set datasets mostly have more nodes than those in the test set, while the number of edges and days are in the same range for both.

We summarize detailed statistics of each token network in  \methodname \  datasets in Table \ref{tab:TGS_all_stats}. In the table, the growth rate is the ratio of label $1$, indicating the increase in the number of edge counts concerning the problem definition defined in Appendix section \ref{appendix:property_prediction}. In addition, we use the novelty and surprise scores introduced by \citet{Poursafaei2022BetterEvaluation}. The novelty score is defined as the average ratio of new edges in each timestamp. The surprise score is defined as the ratio of edges that only appear in the test set. Formally, 

\begin{subequations}
\small
\label{equ:novelty_surprise}
\begin{align}
&novelty = \frac{1}{T} \sum_{t=1}^T \frac{|E^t \setminus E^t_{seen}|}{|E^t|} \tag{\ref{equ:novelty_surprise}a},\label{equ:novelty}\\
& surprise = \frac{|E_{test} \setminus E_{train}|}{|E_{test}|}, \tag{\ref{equ:novelty_surprise}g}\label{equ:surprise}
\end{align}
\end{subequations}

where $E^t$ and $E^t_{seen}$ denotes the set of edges present only in timestamp $t$ and seen in previous timestamps, respectively. $E_{test}$ represents edges that appear in the test set, and edges appearing in the train set are represented as $E_{train}$.

\textbf{Comparison Between Training And Testing Set}.
Nodes, transactions, and length (in days) distribution over the training and testing sets are shown in Figure \ref{fig:TGS_training_testing_characteristic}. Training sets well-support the multi-network model to generalize characteristics of the entire  \methodname \  dataset due to the similarity between nodes, edge and length in days distributions shown in Figures \ref{fig:node_distribution_training}, \ref{fig:edges_distribution_training}, \ref{fig:length_distribution_training} and those distributions across \(84\) token networks of  \methodname \  datasets. In addition, the variance of datasets' characteristics of the testing set is shown in Figures \ref{fig:node_distribution_testing}, \ref{fig:edges_distribution_testing} and \ref{fig:length_distribution_testing}.

\textbf{Node Overlap Analysis.} We analyze the overlap of nodes between different datasets and within each dataset, which helps demonstrate the highly dynamic nature of our datasets. Specifically, we compared the nodes in each test network with those in the training networks and calculated the average overlap. As shown in Table~\ref{tab:node_overlap}, on average, only 2\% of the nodes are common between the training and test datasets, highlighting the rapidly changing structure of these networks.
Furthermore, we analyzed the node overlap within each test dataset by splitting it into the standard train-validation-test setup. We compared the nodes in the 70\% training snapshots with the nodes in the final 15\% test snapshots, and on average, only 4\% of the nodes overlapped. This indicates the highly inductive nature of our model and emphasizes the zero-shot challenge it addresses in this domain. These findings underscore the importance of tackling such dynamic and evolving challenges in temporal graph learning.

\begin{table}[ht]
    \centering
    \caption{Overlapping Nodes Statistics}
\label{tab:node_overlap}
  \resizebox{\linewidth}{!}{%
    \begin{tabular}{l | c | c}
    \toprule 
        Dataset & \shortstack{Average Node in Common \\ vs Train Set of \methodname-64 (± std)} & \shortstack{Train vs Test Snapshots \\ Node in Common} \\ \midrule
        MIR & 0.021 ± 0.019 & 0.007 \\
        DOGE2.0 & 0.026 ± 0.033 & 0.015 \\ 
        MUTE & 0.033 ± 0.020 & 0.045 \\ 
        EVERMOON & 0.023 ± 0.033 & 0.043 \\ 
        DERC & 0.020 ± 0.020 & 0.031 \\ 
        ADX & 0.024 ± 0.020 & 0.018 \\ 
        HOICHI & 0.023 ± 0.013 & 0.053 \\ 
        SDEX & 0.024 ± 0.019 & 0.141 \\ 
        BAG & 0.019 ± 0.017 & 0.107 \\ 
        XCN & 0.016 ± 0.010 & 0.034 \\ 
        ETH2x-FLI & 0.038 ± 0.041 & 0.028 \\ 
        stkAAVE & 0.026 ± 0.027 & 0.057 \\ 
        GLM & 0.014 ± 0.015 & 0.047 \\ 
        QOM & 0.018 ± 0.014 & 0.044 \\ 
        WOJAK & 0.025 ± 0.032 & 0.018 \\ 
        DINO & 0.018 ± 0.014 & 0.049 \\ 
        Metis & 0.020 ± 0.013 & 0.041 \\ 
        REPv2 & 0.016 ± 0.017 & 0.013 \\ 
        TRAC & 0.015 ± 0.016 & 0.031 \\ 
        BEPRO & 0.023 ± 0.022 & 0.021 \\ \bottomrule
    \end{tabular}}
\end{table}

\section{Temporal Graph Learning} \label{appendix:singlemodels}
In this section, we give further details about the temporal graph learning models we used as a baseline for our work.

\textbf{Persistence Forecast} uses data from the previous and current weeks to predict the next week's property. If we observe an increasing trend in the number of transactions in the current week compared to the previous week, we predict a similar increasing trend for the following week. This simple model is based on the assumption that trends in transaction networks can persist over time. Our baseline method has three key aspects. First, we do not use any future information to generate the labels. Second, we compare the current week's transaction count to that of the previous week to determine the trend. Finally, if the current week shows an increase, we predict the same trend for the next week. This straightforward approach provides a basic baseline for comparison against more sophisticated predictive models.

\textbf{HTGN} leverages the power of hyperbolic geometry, which is well-suited for capturing hierarchical structures and complex relationships in temporal networks. HTGN maps the temporal graph into hyperbolic space and utilizes hyperbolic graph neural networks and hyperbolic gated recurrent neural networks to model the evolving dynamics. It incorporates two key modules that are hyperbolic temporal contextual self-attention (HTA) and hyperbolic temporal consistency (HTC)-to ensure that temporal dependencies are effectively captured and that the model is both stable and generalizable across various tasks~\cite{menglin2021HTGN}.

\textbf{GraphPulse} addresses the challenge of learning from nodes and edges with different timestamps, which many existing models struggle with. It combines two key techniques: the Mapper method from topological data analysis to extract clustering information from graph nodes and Recurrent Neural Networks (RNNs) for temporal reasoning. This principled approach helps capture both the structure and dynamics of evolving graphs~\cite{shamsi2024graphpulse}.

\textbf{GCLSTM} combines a Graph Convolutional Network (GCN) and Long Short-Term Memory (LSTM) units to handle both the structural and temporal aspects of evolving networks. The GCN is used to capture the local structural properties of the network at each snapshot, while the LSTM learns the temporal evolution of these snapshots over time~\cite{chen2022GCLSTM}.

\textbf{EvolveGCN} is designed to capture the temporal dynamics of graph-structured data. Instead of relying on static node embeddings, EvolveGCN evolves the parameters of a graph convolutional network (GCN) over time. By using a recurrent neural network (RNN) to adapt the GCN parameters, this model is capable of dynamically adjusting during both training and testing, allowing it to handle evolving graphs, even when node sets vary significantly across different time steps~\cite{Aldo2020EvolveGCN}.

\section{Temporal Graph Property Prediction}\label{appendix:property_prediction}

In this study, we define graph property prediction as the task of predicting a specific graph property. In our case, this involves predicting the growth or shrinkage in the number of transactions in the next snapshot. Specifically, given the current weekly snapshot of a network, the objective is to predict the trend—whether the network will experience growth or shrinkage in transaction volume in the following week. This task has significant applications in the financial domain, as it provides insights into the willingness of investors to engage in a network and whether transaction activity is likely to increase. To ensure consistency, we use the same property prediction setting as GraphPulse \citep{shamsi2024graphpulse}, and the formal definition of the graph property is as follows:

\textbf{Definition.} We define network growth in terms of edge count as the predicted graph property. Let \(\mathcal{G}\) represent a graph, \(t\) a specific time, \(\delta_1\) and \(\delta_2\) time intervals, and \(E(t_1, t_n)\) the multi-set of edges between times \(t_1\) and \(t_n\). The property \(P\) is formally expressed as:

\begin{equation}
\begin{aligned}
    P(\mathcal{G}, t_1, t_n, \delta_1, \delta_2) =
    \begin{cases} 
        1, & \text{if } |E(t_n + \delta_1, t_n + \delta_2)| \\
           & \quad > |E(t_1, t_n)|, \\ 
        0, & \text{otherwise.}
    \end{cases}
\end{aligned}
\end{equation}

Setting \(n = 7\), \(\delta_1 = 1\), and \(\delta_2 = 7\), we establish a practical graph property with a 7-day prediction window. This choice is particularly relevant in financial contexts, such as Ethereum asset networks, where it can guide investment decisions, and in social network infrastructure, like Reddit, where it supports maintenance planning.

\textbf{Insights For Transaction Networks.} The graph growth/shrink property prediction in financial networks forecasts changes in transaction numbers (edge count), revealing trends in network activity. A growth in edge count indicates increased investor engagement, while a shrinkage suggests reduced activity or market hesitation. This property helps guide investment strategies, resource allocation, and risk management by providing insights into the evolving dynamics of transaction networks.

In temporal graphs, property predictions provide valuable insights into the dynamics and behaviors of evolving networks. While this work focuses on specific properties, numerous other characteristics can also be defined in this domain to highlight the significance of temporal graph property predictions. For instance, properties like the temporal global efficiency, temporal-correlation coefficient, and temporal betweenness centrality offer additional perspectives by capturing unique aspects of a graph's temporal evolution. These examples further clarify the importance of studying temporal graph properties and their relevance to understanding complex network dynamics. Below, we formalize these three additional temporal graph properties and explain their relevance and insights which can be used in future works, particularly for transaction networks.

\section{Hyperparameters} \label{ap:hyperparameters}
\textbf{Single Models.} We adopt a $70\%-15\%-15\%$ split ratio for the train, validation, and test, respectively, for each token network, and during each epoch, the training model processes all snapshots in chronological order. We train every single model for a minimum of $100$ and a maximum of $250$ epochs with a learning rate set to $15 \times 10^{-4}$. We apply early stopping based on the AUC results on the validation set, with patience and tolerance set to $20$ and $5 \times 10^{-2}$, respectively. \textcolor{black}{Specifically, in HTGN training, the node embeddings are reset at the end of every epoch. We use Binary Cross-Entropy Loss for performance measurement and Adam \cite{kingma2015Adam} as the optimization algorithm. It is important to note that the graph pooling layer, performance measurement, and optimization algorithm are also shared by the \textcolor{black}{multi-network} model training setup. }

\textbf{Multi-network Models.} While following a similar training approach as in the single model training, we make specific adjustments for the \textcolor{black}{multi-network} model training. We set the number of epochs to $300$ with a learning rate of $10^{-4}$ and a train-validation-test chronological split ratio \textcolor{black}{same as single models.} Early stopping is applied based on the validation loss with a tolerance of $5 \times 10^{-2}$ and the patience is set to $30$. The best model is selected based on the validation AUC and used to predict the unseen test dataset.

\section{Hyperbolic Temporal Graph Network} \label{appendix:HTGN}
Hyperbolic geometry has been increasingly recognized for its ability to achieve state-of-the-art performance in several static graph embedding tasks~\cite{menglin2021HTGN}. HTGN is a recent hyperbolic work that shows strong performance in learning over dynamic graphs in a DTDG manner. The model employs a hyperbolic graph neural network (HGNN) to learn the topological dependencies of the nodes and a hyperbolic-gated recurrent unit (HGRU) to capture the temporal dependencies. Temporal contextual attention (HTA) is also used To prevent recurrent neural networks from only emphasizing the most nearby time and to ensure stability along with generalization of the embedding. In addition, HTGN enables updating the model's state at the test time to incorporate new information, which makes it a good candidate for learning the scaling law of TGNNs. \textcolor{black}{In our  \methodname \  framework, we use the HTGN architecture as part of our multi-network model because it excels in dynamic graph learning through hyperbolic geometry. Its strong performance makes it a valuable addition to our approach.}

Given feature vectors $X^E_t$ of snapshot $t$ in Euclidean space, an HGNN layer first adopts an exponential map to project Euclidean space vectors to hyperbolic space as follows $X^{\mathcal{H}}_t = exp^c{(X^E_t)}$, and then performs aggregation and activation similar to GNN but in a hyperbolic manner, $\tilde{X}_t^{\mathcal{H}}$ = $\mathbf{HGNN}(X_t^{\mathcal{H}})$. To prevent recurrent neural networks from only emphasizing the most nearby time and to ensure stability along with generalization of the embedding, HTGN uses temporal contextual attention (HTA) to generalize the lastest \(w\) hidden states such that $\tilde{H}_{t-1}^{\mathcal{H}}$ = $\mathbf{HTA}(H_{t-w};...;H_{t-1})$ ~\cite{menglin2021HTGN}. HGRU takes the outputs from HGNN, $\tilde{X}_t^{\mathcal{H}}$, and the attentive hidden state, $\tilde{H}_{t-1}^{\mathcal{H}}$, from HTA as input to update gates and memory cells and then provides the latest hidden state as the output, $H_t^\mathcal{H}=\mathbf{HGRU}(\tilde{X}_t^{\mathcal{H}}, \tilde{H}_{t-1}^\mathcal{H})$. To interpret hyperbolic embeddings, \cite{menglin2021HTGN} adopt Poincaré ball model with negative curve $-c$, given $c >0$, coresponds to the Riemannian manifold $(\mathbb{H}^{n,c}) = \{x \in \mathbb{R}^n : c||x||^2 <1 \}$ is an open n-dimensional ball. Given a Euclidean space vector $x_i^E \in \mathbb{R}^d$, we consider it as a point in tangent space $\mathcal{T}_{x'}\mathbb{H}^{d,c}$ and adopt the exponential map to project it into hyperbolic space :
\begin{equation}
    x_i^{\mathcal{H}} = exp_{x'}^c(x_i^E)
\end{equation}

Resulting in $x_i^{\mathcal{H}} \in \mathbb{H}^{d,c}$, which is then served as input to the HGNN layer as follows \cite{menglin2021HTGN}:
\begin{subequations}
    \small
    \label{equ:HGNN}

    \begin{align}
         \mathbf{m}_i^{\mathcal{H}}&=W\otimes^c\mathbf{x}_i^{\mathcal{H}}\oplus^c \mathbf{b},\tag{\ref{equ:HGNN}a}\label{equ:HGNNa}\\
       \tilde{\mathbf{m}}_i^{\mathcal{H}}&=\exp_{\mathbf{x'}}^c(\sum_{j \in \mathcal{N}(i)} \alpha_{ij}\log_\mathbf{\mathbf{x}'}^c(\mathbf{m}_i^{\mathcal{H}}))\tag{\ref{equ:HGNN}b},\label{equ:HGNNb}\\
       \tilde{\mathbf{x}}_i^{\mathcal{H}}&= \exp_\mathbf{x'}^c(\sigma({\log_{\mathbf{x'}}^c}(\tilde{\mathbf{m}}_i^{\mathcal{H}})).\tag{\ref{equ:HGNN}c}\label{equ:HGNNc}
    \end{align}
   
\end{subequations}

where $W$, $b$ are learnable parameters and hyperbolic activation function $\sigma$ achieved by applying logarithmic and exponential mapping. HGNN leverages attention-based aggregation by assigning attention score $\alpha_{ij}$ to indicate the importance of neighbour $j$ to node $i$, computed as followed:
\begin{equation}
\small
\begin{aligned}
\alpha_{ij}&=softmax_{(j\in\mathcal{N}(i))}(s_{ij})=\frac{\exp(s_{ij})}{\sum_{j'\in\mathcal{N}_i} \exp(s_{ij'})}, \\
s_{ij}&=\mathrm{LeakReLU}(a^T[\log_{0}^c(m_i^{l})\|\log_{0}^c(m_j^{l})]),
\end{aligned}
\end{equation}

where $a$ is trainable vector and $||$ denotes concatenation operation.

The output of HGNN, $\tilde{X}_t^\mathcal{H}$, is then used as input to HGRU along with attentive hidden state  $\tilde{H}_{t-1}^\mathcal{H}$ obtained by HTA, which generalize $H_{t-1}$ to lastest $w$ snapshots $\{H_{t-w},...,H_{t-1}\}$ \cite{menglin2021HTGN}. Operations behind HGRU are characterized by the following equation \cite{menglin2021HTGN}:

\begin{subequations}
\small
\label{equ:GRU}
\begin{align}
&X_t^E = \log_{\mathbf{x'}}^c(\tilde{X}_t^\mathcal{H})\tag{\ref{equ:GRU}a},\label{equ:GRUa}\\
&H_{t-1}^E = \log_{\mathbf{x'}}^c(\tilde{H}_{t-1}^\mathcal{H})\tag{\ref{equ:GRU}b},\label{equ:GRUb}\\
& P_t^E = \sigma(W_z X_t^E + U_z H_{t-1}^E) \tag{\ref{equ:GRU}c}\label{equ:GRUc}\\
& R_t^E = \sigma(W_r X_t^E + U_r H_{t-1}^E), \tag{\ref{equ:GRU}d}\label{equ:GRUd}\\
& \tilde{H}_t^E = \tanh(W_h X_t^E + U_h (R_t\odot H_{t-1}^E)), \tag{\ref{equ:GRU}e}\label{equ:GRUe}\\
& H_t^E = (1-P_t^E) \odot \tilde{H}_t^E + P_t^E \odot H_{t-1}^E, \tag{\ref{equ:GRU}f}\label{equ:GRUf}\\
& H_t^\mathcal{H} = \exp_{\mathbf{x'}}^c(H_t^E). \tag{\ref{equ:GRU}g}\label{equ:GRUg}
\end{align}
\end{subequations}
where $W_z, W_r, W_h, U_z, U_r, U_h$ are the trainable weight matrices, $P_t^E$ is the update gate to control the output and $R_t^E$ is the reset gate to balance the input and memory \cite{menglin2021HTGN}.

\begin{figure*}[ht]
    \centering
    \includegraphics[width=\linewidth]{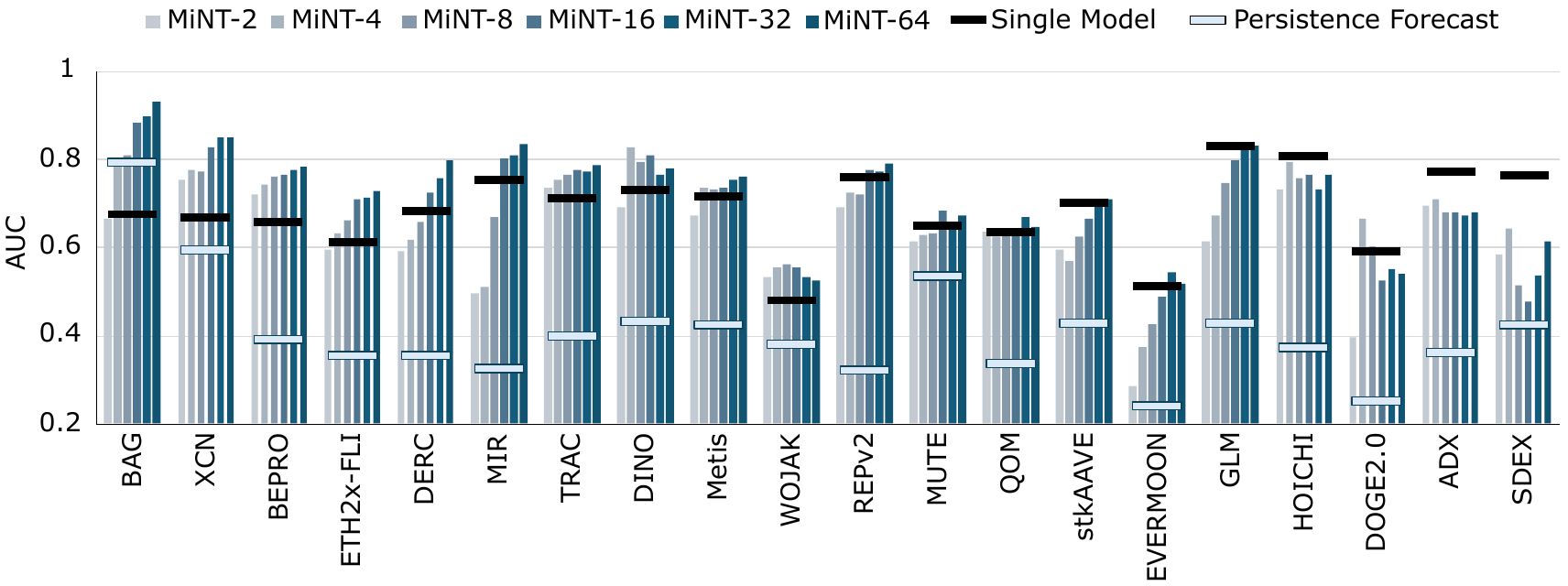}
    \caption{Test AUC of multi-network models trained on $2^n$ datasets where $n\in [1, 6]$ and evaluated on unseen test datasets. Comparing the performance with single models trained and tested on each dataset and persistence forecast results.}
    \label{fig:htgn-all}

\end{figure*}

\section{Additional Results} \label{app:detailed_results}

Here, we present the test results for the six multi-network models trained on different network sizes, as well as the single model and persistence forecast results. Figure~\ref{fig:htgn-all} illustrates the AUC of these models on the test set. In most datasets, multi-network models outperform the single model, and in all datasets, they outperform the persistence forecast.
We have also compared our model against additional state-of-the-art models, specifically including EvolveGCN~\cite{Aldo2020EvolveGCN}, GC-LSTM~\cite{chen2022GCLSTM} and the only model designed for temporal graph properties prediction, GraphPulse~\cite{shamsi2024graphpulse} as baselines for the test set. In Table~\ref{tab:allmodel-auc} and Table~\ref{detailed_results-AP} the average and standard deviation of AUC and AP are presented respectively for all models. Surprisingly, \methodname-64 stands out as the best model, consistently achieving competitive performance in a greater number of datasets for both AUC and AP scores compared to all other models. Similarly, MN-32 shows strong performance, attaining the highest score in several datasets and placing second in numerous others, but it does not surpass MN-64 in overall rankings. These results show the power of multi-network models in performing downstream tasks on unseen datasets.  Importantly, this high level of performance is achieved through zero-shot inference, meaning that the model was not specifically trained on these datasets. In contrast, other models, including GraphPulse, were trained directly for the datasets they evaluated. This considerable difference underscores the potential of \methodname-64 and highlights the power of zero-shot learning in effectively leveraging knowledge across different temporal graphs.

Table~\ref{detailed_test_results_gclstm} presents the detailed performance of all \methodname models trained with GCLSTM. Notably, we observed a consistent trend with GCLSTM: as the model was trained on a larger number of networks, its zero-shot inference performance improved significantly. This highlights the positive impact of training on diverse networks for enhancing the model's generalization capabilities.

We further compare the performance between \methodname-64 and single models on the training datasets shown in Table \ref{tab:train_set}. There are \(38\) token networks \methodname-64 that outperform single models, while single models achieve better results in \(26\) datasets. In addition, the improvement in AUCs of \methodname-64 compared to single models across all training datasets is \(2\%\) on average, reflecting the improvement obtained from training multiple networks.

\begin{table*}[ht]\label{tab:detailed_results}
    \centering
 \caption{\textcolor{black}{\textbf{AUC} scores of multi-network models, single models, and persistence forecasts on test sets across three seeds, including comparisons with state-of-the-art models EvolveGCN, GC-LSTM and GraphPulse. The best performance is shown in bold, and the second best is underlined.}}
  \label{tab:allmodel-auc}

  \resizebox{\linewidth}{!}{%
      \begin{tabular}{l | c | c | c |c | c | c |c | c| c| c | c}
    \toprule 
  
      Token  & Per. Fore. & GraphPulse & HTGN & GCLSTM & EvolveGCN & \methodname-2 & \methodname-4 & \methodname-8 & \methodname-16 & \methodname-32 & \methodname-64 \\ \midrule

        WOJAK & 0.378 & 0.467$\pm$ \scriptsize0.030 & 0.479 $\pm$ \scriptsize0.005 & 0.484 $\pm$ \scriptsize0.000 & 0.505 $\pm$ \scriptsize0.023 & 0.534 $\pm$ \scriptsize0.020 & \underline{0.556 $\pm$ \scriptsize0.029} & \textbf{0.561 $\pm$ \scriptsize0.018} & 0.556 $\pm$ \scriptsize0.016 & 0.534 $\pm$ \scriptsize0.017 & 0.524 $\pm$ \scriptsize0.027  \\
        
        DOGE2.0 & 0.250 & 0.384  $\pm$ \scriptsize0.18 & 0.590 $\pm$ \scriptsize0.059 & 0.538 $\pm$ \scriptsize0.000 & 0.551 $\pm$ \scriptsize0.022 & 0.397 $\pm$ \scriptsize0.124 & \textbf{0.667 $\pm$ \scriptsize0.219} & \underline{0.603 $\pm$ \scriptsize0.080} & 0.526 $\pm$ \scriptsize0.059 & 0.551 $\pm$ \scriptsize0.022 & 0.538 $\pm$ \scriptsize0.038  \\
        
        EVERMOON & 0.241 & {0.519 $\pm$ \scriptsize0.130} & 0.512 $\pm$ \scriptsize0.023 & \textbf{0.562 $\pm$ \scriptsize0.179} & 0.451 $\pm$ \scriptsize0.046 & 0.287 $\pm$ \scriptsize0.153 & 0.373 $\pm$ \scriptsize0.037 & 0.426 $\pm$ \scriptsize0.065 & 0.488 $\pm$ \scriptsize0.054 & \underline{0.543 $\pm$ \scriptsize0.075} & 0.517 $\pm$ \scriptsize0.039  \\
        
        QOM & 0.334 & \textbf{0.775 $\pm$ \scriptsize0.011} & 0.633 $\pm$ \scriptsize0.017 & 0.612 $\pm$ \scriptsize0.001 & 0.618 $\pm$ \scriptsize0.002 & 0.635 $\pm$ \scriptsize0.061 & 0.624 $\pm$ \scriptsize0.025 & 0.633 $\pm$ \scriptsize0.032 & 0.644 $\pm$ \scriptsize0.009 & \underline{0.669 $\pm$ \scriptsize0.034} & {0.647 $\pm$ \scriptsize0.019}  \\
        
        SDEX & 0.423 &0.436 $\pm$ \scriptsize0.030& \textbf{0.762 $\pm$ \scriptsize0.034} & 0.720 $\pm$ \scriptsize0.002 & \underline{0.733 $\pm$ \scriptsize0.028} & 0.585 $\pm$ \scriptsize0.139 & 0.643 $\pm$ \scriptsize0.021 & 0.515 $\pm$ \scriptsize0.031 & 0.476 $\pm$ \scriptsize0.010 & 0.536 $\pm$ \scriptsize0.042 & 0.614 $\pm$ \scriptsize0.020  \\
        
        ETH2x-FLI & 0.355 & {0.666 $\pm$ \scriptsize0.047} & 0.610 $\pm$ \scriptsize0.059 & 0.670 $\pm$ \scriptsize0.009 & 0.688 $\pm$ \scriptsize0.010 & 0.595 $\pm$ \scriptsize0.083 & 0.632 $\pm$ \scriptsize0.019 & 0.663 $\pm$ \scriptsize0.018 & 0.710 $\pm$ \scriptsize0.037 & \underline{0.715 $\pm$ \scriptsize0.032} & \textbf{0.729 $\pm$ \scriptsize0.015}  \\
        
        BEPRO & 0.393 & \textbf{0.783 $\pm$ \scriptsize0.003} & 0.655 $\pm$ \scriptsize0.038 & 0.632 $\pm$ \scriptsize0.019 & 0.610 $\pm$ \scriptsize0.012 & 0.720 $\pm$ \scriptsize0.028 & 0.742 $\pm$ \scriptsize0.013 & 0.762 $\pm$ \scriptsize0.007 & 0.765 $\pm$ \scriptsize0.024 & {0.776 $\pm$ \scriptsize0.008} & \underline{0.782 $\pm$ \scriptsize0.003}  \\
        
        XCN & 0.592 & 0.821 $\pm$ \scriptsize0.004  & 0.668 $\pm$ \scriptsize0.099 & 0.306 $\pm$ \scriptsize0.092 & 0.512 $\pm$ \scriptsize0.067 & 0.754 $\pm$ \scriptsize0.025 & 0.774 $\pm$ \scriptsize0.062 & 0.773 $\pm$ \scriptsize0.076 & 0.827 $\pm$ \scriptsize0.061 & \underline{0.848 $\pm$ \scriptsize0.000} & \textbf{0.851 $\pm$ \scriptsize0.043} \\
        
        BAG & 0.792 & \textbf{0.934 $\pm$ \scriptsize0.020} & 0.673 $\pm$ \scriptsize0.227 & 0.196 $\pm$ \scriptsize0.179 & 0.329 $\pm$ \scriptsize0.040 & 0.667 $\pm$ \scriptsize0.134 & 0.802 $\pm$ \scriptsize0.155 & 0.808 $\pm$ \scriptsize0.095 & 0.884 $\pm$ \scriptsize0.044 & {0.898 $\pm$ \scriptsize0.075} & \underline{0.931 $\pm$ \scriptsize0.028}  \\
        
        TRAC & 0.400 & 0.767 $\pm$ \scriptsize0.001 & 0.712 $\pm$ \scriptsize0.071 & 0.748 $\pm$ \scriptsize0.000 & 0.748 $\pm$ \scriptsize0.000 & 0.734 $\pm$ \scriptsize0.012 & 0.752 $\pm$ \scriptsize0.009 & 0.764 $\pm$ \scriptsize0.012 & 0.776 $\pm$ \scriptsize0.012 & \underline{0.770 $\pm$ \scriptsize0.007} & \textbf{0.785 $\pm$ \scriptsize0.008}  \\
        
        DERC & 0.353 & \underline{0.769 $\pm$ \scriptsize0.040} & 0.683 $\pm$ \scriptsize0.013 & 0.703 $\pm$ \scriptsize0.022 & 0.669 $\pm$ \scriptsize0.009 & 0.593 $\pm$ \scriptsize0.108 & 0.617 $\pm$ \scriptsize0.030 & 0.657 $\pm$ \scriptsize0.009 & 0.723 $\pm$ \scriptsize0.058 & {0.756 $\pm$ \scriptsize0.045} & \textbf{0.798 $\pm$ \scriptsize0.027}  \\
        
        Metis & 0.423 & \textbf{0.812 $\pm$ \scriptsize0.011} & 0.715 $\pm$ \scriptsize0.122 & 0.646 $\pm$ \scriptsize0.023 & 0.688 $\pm$ \scriptsize0.027 & 0.672 $\pm$ \scriptsize0.103 & 0.734 $\pm$ \scriptsize0.017 & 0.730 $\pm$ \scriptsize0.036 & 0.734 $\pm$ \scriptsize0.016 & {0.753 $\pm$ \scriptsize0.005} & \underline{0.760 $\pm$ \scriptsize0.025}  \\
        
        REPv2 & 0.321 & \textbf{0.830 $\pm$ \scriptsize0.001} & 0.760 $\pm$ \scriptsize0.012 & 0.725 $\pm$ \scriptsize0.014 & 0.709 $\pm$ \scriptsize0.002 & 0.690 $\pm$ \scriptsize0.024 & 0.725 $\pm$ \scriptsize0.023 & 0.719 $\pm$ \scriptsize0.022 & 0.774 $\pm$ \scriptsize0.013 & {0.773 $\pm$ \scriptsize0.013} & \underline{0.789 $\pm$ \scriptsize0.020}  \\
        
        DINO & 0.431 & {0.801 $\pm$ \scriptsize0.020} &0.730 $\pm$ \scriptsize0.195 & \textbf{0.874 $\pm$ \scriptsize0.028} & \underline{0.868 $\pm$ \scriptsize0.029} & 0.692 $\pm$ \scriptsize0.140 & 0.827 $\pm$ \scriptsize0.112 & 0.794 $\pm$ \scriptsize0.096 & 0.809 $\pm$ \scriptsize0.087 & 0.764 $\pm$ \scriptsize0.048 & 0.779 $\pm$ \scriptsize0.113  \\
        
        HOICHI & 0.374 & 0.714 $\pm$ \scriptsize0.010& 0.807 $\pm$ \scriptsize0.047 & \textbf{0.857 $\pm$ \scriptsize0.000} & \underline{0.856 $\pm$ \scriptsize0.001} & 0.733 $\pm$ \scriptsize0.101 & 0.795 $\pm$ \scriptsize0.025 & 0.759 $\pm$ \scriptsize0.040 & 0.763 $\pm$ \scriptsize0.026 & 0.731 $\pm$ \scriptsize0.029 & 0.765 $\pm$ \scriptsize0.018  \\
        
        MUTE & 0.536 & \textbf{0.779 $\pm$ \scriptsize0.004} & 0.649 $\pm$ \scriptsize0.015 & 0.593 $\pm$ \scriptsize0.030 & 0.617 $\pm$ \scriptsize0.010 & 0.613 $\pm$ \scriptsize0.027 & 0.627 $\pm$ \scriptsize0.024 & 0.633 $\pm$ \scriptsize0.024 & \underline{0.684 $\pm$ \scriptsize0.042} & 0.657 $\pm$ \scriptsize0.035 & {0.673 $\pm$ \scriptsize0.013}  \\
        
        GLM & 0.427 & 0.769 $\pm$ \scriptsize0.018& \underline{0.830 $\pm$ \scriptsize0.029} & 0.451 $\pm$ \scriptsize0.003 & 0.501 $\pm$ \scriptsize0.033 & 0.613 $\pm$ \scriptsize0.115 & 0.671 $\pm$ \scriptsize0.034 & 0.746 $\pm$ \scriptsize0.082 & 0.800 $\pm$ \scriptsize0.062 & 0.826 $\pm$ \scriptsize0.035 & \textbf{0.831 $\pm$ \scriptsize0.024} \\
        
        MIR & 0.327 & {0.689 $\pm$ \scriptsize0.097} & 0.750 $\pm$ \scriptsize0.005 & 0.768 $\pm$ \scriptsize0.026 & 0.745 $\pm$ \scriptsize0.015 & 0.497 $\pm$ \scriptsize0.192 & 0.510 $\pm$ \scriptsize0.015 & 0.669 $\pm$ \scriptsize0.103 & 0.800 $\pm$ \scriptsize0.044 & \underline{0.809 $\pm$ \scriptsize0.022} & \textbf{0.836 $\pm$ \scriptsize0.016}  \\
        
        stkAAVE & 0.426 & \textbf{0.743 $\pm$ \scriptsize0.006} & {0.702 $\pm$ \scriptsize0.042} & 0.368 $\pm$ \scriptsize0.011 & 0.397 $\pm$ \scriptsize0.022 & 0.597 $\pm$ \scriptsize0.076 & 0.571 $\pm$ \scriptsize0.026 & 0.626 $\pm$ \scriptsize0.023 & 0.666 $\pm$ \scriptsize0.033 & 0.696 $\pm$ \scriptsize0.027 & \underline{0.709 $\pm$ \scriptsize0.022}  \\
        
        ADX & 0.362 & \textbf{0.784 $\pm$ \scriptsize0.002} & \underline{0.769 $\pm$ \scriptsize0.018} & {0.723 $\pm$ \scriptsize0.002} & 0.718 $\pm$ \scriptsize0.004 & 0.695 $\pm$ \scriptsize0.003 & 0.708 $\pm$ \scriptsize0.025 & 0.680 $\pm$ \scriptsize0.008 & 0.678 $\pm$ \scriptsize0.019 & 0.671 $\pm$ \scriptsize0.015 & 0.679 $\pm$ \scriptsize0.024  \\

\bottomrule
   
    \end{tabular}%
    }
\end{table*}

\begin{table*}[ht]
    \centering
 \caption{\textcolor{black}{\textbf{AP} scores of multi-network models, single models, and persistence forecasts on test sets across three seeds, including comparisons with state-of-the-art models EvolveGCN, GC-LSTM and GraphPulse. The best performance is shown in bold, and the second best is underlined.}}
\label{detailed_results-AP}
  \resizebox{\linewidth}{!}{%
      \begin{tabular}{l | c | c | c |c | c | c |c | c| c| c |c}
    \toprule 
  
      Token  & Per. Fore. & GraphPulse & HTGN & GCLSTM & EvolveGCN & \methodname-2 & \methodname-4 & \methodname-8 & \methodname-16 & \methodname-32 & \methodname-64   \\ \midrule

        WOJAK & 0.658 & \textbf{0.863$\pm$ \scriptsize0.006} & 0.812 $\pm$ \scriptsize0.003 & 0.812 $\pm$ \scriptsize0.000 & 0.827 $\pm$ \scriptsize0.017 & 0.832 $\pm$ \scriptsize0.009 & 0.836 $\pm$ \scriptsize0.015 & 0.842 $\pm$ \scriptsize0.015 & \underline{0.850 $\pm$ \scriptsize0.006} & 0.842 $\pm$ \scriptsize0.008 & {0.837 $\pm$ \scriptsize0.019}  \\
        
        DOGE2.0 & 0.2 & \textbf{0.966  $\pm$ \scriptsize0.002} &0.933 $\pm$ \scriptsize0.010 & 0.925 $\pm$ \scriptsize0.000 & 0.927 $\pm$ \scriptsize0.004 & 0.889 $\pm$ \scriptsize0.031 & \underline{0.940 $\pm$ \scriptsize0.050} & {0.936 $\pm$ \scriptsize0.014} & 0.920 $\pm$ \scriptsize0.014 & 0.927 $\pm$ \scriptsize0.004 & 0.921 $\pm$ \scriptsize0.014  \\
        
        EVERMOON & 0.469 & \textbf{0.768 $\pm$ \scriptsize0.01} & {0.585 $\pm$ \scriptsize0.065} & \underline{0.612 $\pm$ \scriptsize0.200} & 0.494 $\pm$ \scriptsize0.017 & 0.442 $\pm$ \scriptsize0.059 & 0.508 $\pm$ \scriptsize0.045 & 0.542 $\pm$ \scriptsize0.031 & 0.530 $\pm$ \scriptsize0.040 & 0.567 $\pm$ \scriptsize0.053 & 0.551 $\pm$ \scriptsize0.021  \\
        
        QOM & 0.315 & {0.840 $\pm$ \scriptsize0.002} & 0.623 $\pm$ \scriptsize0.024 & 0.592 $\pm$ \scriptsize0.001 & 0.597 $\pm$ \scriptsize0.002 & 0.632 $\pm$ \scriptsize0.070 & 0.617 $\pm$ \scriptsize0.022 & 0.616 $\pm$ \scriptsize0.007 & 0.626 $\pm$ \scriptsize0.020 & \textbf{0.648 $\pm$ \scriptsize0.027} & \underline{0.635 $\pm$ \scriptsize0.027}  \\
        
        SDEX & 0.212 & 0.662 $\pm$ \scriptsize0.017 & \textbf{0.825 $\pm$ \scriptsize0.048} & 0.725 $\pm$ \scriptsize0.002 & \underline{0.750 $\pm$ \scriptsize0.025} & 0.723 $\pm$ \scriptsize0.039 & 0.725 $\pm$ \scriptsize0.021 & 0.650 $\pm$ \scriptsize0.046 & 0.628 $\pm$ \scriptsize0.036 & 0.697 $\pm$ \scriptsize0.064 & 0.699 $\pm$ \scriptsize0.021  \\
        
        ETH2x-FLI & 0.381 & \textbf{0.836 $\pm$ \scriptsize0.015} & 0.590 $\pm$ \scriptsize0.103 & 0.735 $\pm$ \scriptsize0.018 & {0.756 $\pm$ \scriptsize0.013} & 0.607 $\pm$ \scriptsize0.122 & 0.621 $\pm$ \scriptsize0.039 & 0.658 $\pm$ \scriptsize0.057 & 0.745 $\pm$ \scriptsize0.051 & 0.737 $\pm$ \scriptsize0.049 & \underline{0.784 $\pm$ \scriptsize0.007}  \\
        
        BEPRO & 0.374 & {0.802 $\pm$ \scriptsize0.001} & 0.686 $\pm$ \scriptsize0.042 & 0.637 $\pm$ \scriptsize0.022 & 0.622 $\pm$ \scriptsize0.009 & 0.743 $\pm$ \scriptsize0.033 & 0.769 $\pm$ \scriptsize0.015 & 0.799 $\pm$ \scriptsize0.016 & 0.804 $\pm$ \scriptsize0.034 & \underline{0.815 $\pm$ \scriptsize0.007} & \textbf{0.816 $\pm$ \scriptsize0.014}  \\
        
        XCN & 0.413 & 0.793 $\pm$ \scriptsize0.002  & 0.687 $\pm$ \scriptsize0.085 & 0.420 $\pm$ \scriptsize0.032 & 0.555 $\pm$ \scriptsize0.073 & 0.708 $\pm$ \scriptsize0.065 & 0.765 $\pm$ \scriptsize0.080 & 0.781 $\pm$ \scriptsize0.082 & 0.829 $\pm$ \scriptsize0.057 & \underline{0.851 $\pm$ \scriptsize0.023} & \textbf{0.861 $\pm$ \scriptsize0.042}  \\
        
        BAG & 0.504 & \textbf{0.957 $\pm$ \scriptsize0.004} & 0.523 $\pm$ \scriptsize0.290 & 0.235 $\pm$ \scriptsize0.041 & 0.263 $\pm$ \scriptsize0.011 & 0.474 $\pm$ \scriptsize0.152 & 0.699 $\pm$ \scriptsize0.193 & 0.682 $\pm$ \scriptsize0.160 & 0.784 $\pm$ \scriptsize0.118 & {0.829 $\pm$ \scriptsize0.119} & \underline{0.889 $\pm$ \scriptsize0.043}  \\
        
        TRAC & 0.4 & \textbf{0.767 $\pm$ \scriptsize0.002} & 0.685 $\pm$ \scriptsize0.074 & 0.716 $\pm$ \scriptsize0.006 & 0.722 $\pm$ \scriptsize0.001 & 0.705 $\pm$ \scriptsize0.013 & 0.734 $\pm$ \scriptsize0.012 & 0.741 $\pm$ \scriptsize0.006 & \underline{0.764 $\pm$ \scriptsize0.015} & 0.741 $\pm$ \scriptsize0.015 & {0.758 $\pm$ \scriptsize0.021}  \\
        
        DERC & 0.39 & \textbf{0.773 $\pm$ \scriptsize0.004}  & 0.532 $\pm$ \scriptsize0.021 & 0.621 $\pm$ \scriptsize0.053 & 0.513 $\pm$ \scriptsize0.012 & 0.505 $\pm$ \scriptsize0.157 & 0.477 $\pm$ \scriptsize0.021 & 0.516 $\pm$ \scriptsize0.030 & 0.639 $\pm$ \scriptsize0.118 & {0.700 $\pm$ \scriptsize0.080} & \underline{0.741 $\pm$ \scriptsize0.024}  \\
        
        Metis & 0.38 & \textbf{0.801 $\pm$ \scriptsize0.003} & 0.601 $\pm$ \scriptsize0.187 & 0.575 $\pm$ \scriptsize0.041 & 0.577 $\pm$ \scriptsize0.006 & 0.532 $\pm$ \scriptsize0.126 & {0.645 $\pm$ \scriptsize0.029} & 0.632 $\pm$ \scriptsize0.056 & 0.611 $\pm$ \scriptsize0.021 & \underline{0.647 $\pm$ \scriptsize0.026} & 0.639 $\pm$ \scriptsize0.077  \\
        
        REPv2 & 0.376 & \textbf{0.797 $\pm$ \scriptsize0.003} & \underline{0.758 $\pm$ \scriptsize0.033} & 0.691 $\pm$ \scriptsize0.006 & 0.689 $\pm$ \scriptsize0.001 & 0.610 $\pm$ \scriptsize0.063 & 0.619 $\pm$ \scriptsize0.019 & 0.635 $\pm$ \scriptsize0.042 & 0.705 $\pm$ \scriptsize0.027 & 0.721 $\pm$ \scriptsize0.004 & {0.729 $\pm$ \scriptsize0.011}  \\
        
        DINO & 0.480 & {0.871 $\pm$ \scriptsize0.026}  &  0.747 $\pm$ \scriptsize0.175 & \textbf{0.881 $\pm$ \scriptsize0.029} & \underline{0.875 $\pm$ \scriptsize0.024} & 0.738 $\pm$ \scriptsize0.113 & 0.842 $\pm$ \scriptsize0.102 & 0.793 $\pm$ \scriptsize0.094 & 0.824 $\pm$ \scriptsize0.077 & 0.753 $\pm$ \scriptsize0.030 & 0.765 $\pm$ \scriptsize0.119  \\
        
        HOICHI & 0.602 & {0.623 $\pm$ \scriptsize0.003}  & \underline{0.666 $\pm$ \scriptsize0.062} & 0.650 $\pm$ \scriptsize0.000 & 0.658 $\pm$ \scriptsize0.011 & 0.531 $\pm$ \scriptsize0.109 & \textbf{0.677 $\pm$ \scriptsize0.049} & 0.605 $\pm$ \scriptsize0.037 & 0.609 $\pm$ \scriptsize0.016 & 0.551 $\pm$ \scriptsize0.045 & 0.594 $\pm$ \scriptsize0.012  \\
        
        MUTE & 0.38 & \textbf{0.726 $\pm$ \scriptsize0.002} & 0.615 $\pm$ \scriptsize0.049 & 0.504 $\pm$ \scriptsize0.012 & 0.527 $\pm$ \scriptsize0.015 & 0.579 $\pm$ \scriptsize0.023 & 0.612 $\pm$ \scriptsize0.041 & 0.603 $\pm$ \scriptsize0.058 & \underline{0.675 $\pm$ \scriptsize0.032} & 0.609 $\pm$ \scriptsize0.021 & {0.647 $\pm$ \scriptsize0.048}  \\
        
        GLM & 0.387 & 0.712 $\pm$ \scriptsize0.047 & 0.797 $\pm$ \scriptsize0.024 & 0.513 $\pm$ \scriptsize0.001 & 0.529 $\pm$ \scriptsize0.013 & 0.598 $\pm$ \scriptsize0.123 & 0.651 $\pm$ \scriptsize0.031 & 0.709 $\pm$ \scriptsize0.088 & 0.783 $\pm$ \scriptsize0.092 & \underline{0.819 $\pm$ \scriptsize0.035} & \textbf{0.838 $\pm$ \scriptsize0.032}  \\
        
        MIR & 0.405 & {0.766 $\pm$ \scriptsize0.041} & 0.751 $\pm$ \scriptsize0.003 & 0.765 $\pm$ \scriptsize0.012 & 0.752 $\pm$ \scriptsize0.007 & 0.493 $\pm$ \scriptsize0.212 & 0.442 $\pm$ \scriptsize0.024 & 0.645 $\pm$ \scriptsize0.133 & 0.783 $\pm$ \scriptsize0.064 & \underline{0.799 $\pm$ \scriptsize0.015} & \textbf{0.811 $\pm$ \scriptsize0.019}  \\
        stkAAVE & 0.207 & \underline{0.751 $\pm$ \scriptsize0.005} & {0.750 $\pm$ \scriptsize0.020} & 0.506 $\pm$ \scriptsize0.003 & 0.493 $\pm$ \scriptsize0.009 & 0.662 $\pm$ \scriptsize0.066 & 0.622 $\pm$ \scriptsize0.011 & 0.694 $\pm$ \scriptsize0.021 & 0.730 $\pm$ \scriptsize0.037 & 0.741 $\pm$ \scriptsize0.020 & \textbf{0.759 $\pm$ \scriptsize0.019}  \\
        
        ADX & 0.372 & \textbf{0.765 $\pm$ \scriptsize0.003} & \underline{0.758 $\pm$ \scriptsize0.017} & 0.666 $\pm$ \scriptsize0.002 & 0.661 $\pm$ \scriptsize0.017 & 0.638 $\pm$ \scriptsize0.021 & {0.667 $\pm$ \scriptsize0.040} & 0.632 $\pm$ \scriptsize0.010 & 0.621 $\pm$ \scriptsize0.013 & 0.622 $\pm$ \scriptsize0.018 & 0.628 $\pm$ \scriptsize0.012  \\
      
\bottomrule
   
    \end{tabular}%
    }
\end{table*}

\begin{table*}[ht]
    \centering
 \caption{\textcolor{black}{\textbf{AP and AUC} scores of GCLSTM-based multi-network models on test sets across three seeds. The best performance is shown in bold, and the second best is underlined.}}
\label{detailed_test_results_gclstm}
  \resizebox{\linewidth}{!}{%
      \begin{tabular}{l | cccccc|cccccc}
    \toprule 

    & \multicolumn{6}{c}{\textbf{AUC}}  &  \multicolumn{6}{c}{\textbf{AP}} \\
  
      Token  &  \methodname-2 & \methodname-4 & \methodname-8 & \methodname-16 & \methodname-32 & \methodname-64  & \methodname-2 & \methodname-4 & \methodname-8 & \methodname-16 & \methodname-32 & \methodname-64 \\ \midrule

    MIR & 0.653 $\pm$ \scriptsize0.154 & 0.638 $\pm$ \scriptsize0.090 & 0.588 $\pm$ \scriptsize0.135 & \underline{0.765 $\pm$ \scriptsize0.049} & 0.742 $\pm$ \scriptsize0.036 & \textbf{0.789 $\pm$ \scriptsize0.016} & 0.667 $\pm$ \scriptsize0.153 & 0.602 $\pm$ \scriptsize0.134 & 0.550 $\pm$ \scriptsize0.166 & 0.750 $\pm$ \scriptsize0.019 & \underline{0.758 $\pm$ \scriptsize0.016} & \textbf{0.777 $\pm$ \scriptsize0.013} \\
    
    DOGE2 & 0.487 $\pm$ \scriptsize0.089 & \underline{0.590 $\pm$ \scriptsize0.146} & 0.487 $\pm$ \scriptsize0.219 & 0.282 $\pm$ \scriptsize0.097 & \textbf{0.769 $\pm$ \scriptsize0.133} & 0.551 $\pm$ \scriptsize0.022 & 0.910 $\pm$ \scriptsize0.019 & \underline{0.930 $\pm$ \scriptsize0.030} & 0.907 $\pm$ \scriptsize0.046 & 0.839 $\pm$ \scriptsize0.057 & \textbf{0.965 $\pm$ \scriptsize0.024} & 0.927 $\pm$ \scriptsize0.004 \\
    
    MUTE & 0.592 $\pm$ \scriptsize0.076 & 0.627 $\pm$ \scriptsize0.018 & 0.561 $\pm$ \scriptsize0.035 & 0.589 $\pm$ \scriptsize0.009 & \underline{0.627 $\pm$ \scriptsize0.009} & \textbf{0.636 $\pm$ \scriptsize0.003} & 0.534 $\pm$ \scriptsize0.056 & 0.555 $\pm$ \scriptsize0.017 & 0.502 $\pm$ \scriptsize0.022 & 0.501 $\pm$ \scriptsize0.006 & \underline{0.563 $\pm$ \scriptsize0.002} & \textbf{0.568 $\pm$ \scriptsize0.002} \\
    
    EVERMOON & 0.429 $\pm$ \scriptsize0.078 & 0.318 $\pm$ \scriptsize0.152 & 0.306 $\pm$ \scriptsize0.085 & 0.315 $\pm$ \scriptsize0.154 & \underline{0.420 $\pm$ \scriptsize0.084} & \textbf{0.494 $\pm$ \scriptsize0.048} & 0.493 $\pm$ \scriptsize0.095 & 0.423 $\pm$ \scriptsize0.097 & 0.427 $\pm$ \scriptsize0.037 & 0.447 $\pm$ \scriptsize0.123 & \underline{0.530 $\pm$ \scriptsize0.048} & \textbf{0.560 $\pm$ \scriptsize0.010} \\
    
    DERC & 0.614 $\pm$ \scriptsize0.129 & 0.618 $\pm$ \scriptsize0.058 & 0.569 $\pm$ \scriptsize0.085 & \textbf{0.736 $\pm$ \scriptsize0.027} & 0.647 $\pm$ \scriptsize0.054 & \underline{0.696 $\pm$ \scriptsize0.011} & 0.541 $\pm$ \scriptsize0.150 & 0.546 $\pm$ \scriptsize0.113 & 0.460 $\pm$ \scriptsize0.078 & \textbf{0.693 $\pm$ \scriptsize0.032} & 0.559 $\pm$ \scriptsize0.087 & \underline{0.629 $\pm$ \scriptsize0.012 }\\
    
    ADX &\textbf{ 0.692 $\pm$ \scriptsize0.007} & 0.605 $\pm$ \scriptsize0.182 & 0.674 $\pm$ \scriptsize0.008 & 0.676 $\pm$ \scriptsize0.003 & \underline{0.678 $\pm$ \scriptsize0.004} & 0.674 $\pm$ \scriptsize0.022 & 0.614 $\pm$ \scriptsize0.011 & 0.583 $\pm$ \scriptsize0.147 & \textbf{0.634 $\pm$ \scriptsize0.024} & 0.609 $\pm$ \scriptsize0.005 & \underline{0.617 $\pm$ \scriptsize0.004} & 0.611 $\pm$ \scriptsize0.010 \\
    
    HOICHI & 0.663 $\pm$ \scriptsize0.312 & 0.793 $\pm$ \scriptsize0.065 & 0.633 $\pm$ \scriptsize0.197 & \underline{0.817 $\pm$ \scriptsize0.010} & 0.816 $\pm$ \scriptsize0.043 & \textbf{0.847 $\pm$ \scriptsize0.005} & 0.529 $\pm$ \scriptsize0.240 & 0.602 $\pm$ \scriptsize0.066 & 0.471 $\pm$ \scriptsize0.178 & \underline{0.637 $\pm$ \scriptsize0.016} & 0.630 $\pm$ \scriptsize0.055 & \textbf{0.656 $\pm$ \scriptsize0.014} \\
    
    SDEX & 0.619 $\pm$ \scriptsize0.210 & 0.721 $\pm$ \scriptsize0.032 & 0.574 $\pm$ \scriptsize0.233 & \textbf{0.741 $\pm$ \scriptsize0.014} & 0.717 $\pm$ \scriptsize0.020 & \underline{0.724 $\pm$ \scriptsize0.002} & 0.678 $\pm$ \scriptsize0.115 & \underline{0.732 $\pm$ \scriptsize0.024} & 0.670 $\pm$ \scriptsize0.092 & \textbf{0.752 $\pm$ \scriptsize0.007} & 0.728 $\pm$ \scriptsize0.009 & 0.729 $\pm$ \scriptsize0.002 \\
    
    BAG & \textbf{0.573 $\pm$ \scriptsize0.072} & 0.525 $\pm$ \scriptsize0.010 & 0.374 $\pm$ \scriptsize0.029 & 0.442 $\pm$ \scriptsize0.039 & 0.469 $\pm$ \scriptsize0.060 & \underline{0.529 $\pm$ \scriptsize0.023} & \textbf{0.358 $\pm$ \scriptsize0.036} & 0.334 $\pm$ \scriptsize0.005 & 0.277 $\pm$ \scriptsize0.010 & 0.303 $\pm$ \scriptsize0.013 & 0.311 $\pm$ \scriptsize0.025 & \underline{0.337 $\pm$ \scriptsize0.009 }\\
    
    XCN & \textbf{0.753 $\pm$ \scriptsize0.026} & \underline{0.739 $\pm$ \scriptsize0.005} & 0.726 $\pm$ \scriptsize0.014 & 0.736 $\pm$ \scriptsize0.006 & 0.731 $\pm$ \scriptsize0.005 & 0.733 $\pm$ \scriptsize0.003 & \textbf{0.690 $\pm$ \scriptsize0.064} & 0.657 $\pm$ \scriptsize0.009 & \underline{0.665 $\pm$ \scriptsize0.031} & 0.656 $\pm$ \scriptsize0.007 & 0.650 $\pm$ \scriptsize0.003 & 0.653 $\pm$ \scriptsize0.002 \\
    
    ETH2x-FLI & 0.621 $\pm$ \scriptsize0.119 & 0.615 $\pm$ \scriptsize0.074 & 0.542 $\pm$ \scriptsize0.086 & \underline{0.675 $\pm$ \scriptsize0.008} & 0.666 $\pm$ \scriptsize0.021 & \textbf{0.697 $\pm$ \scriptsize0.010} & 0.669 $\pm$ \scriptsize0.165 & 0.669 $\pm$ \scriptsize0.084 & 0.570 $\pm$ \scriptsize0.154 & \underline{0.752 $\pm$ \scriptsize0.015} & 0.747 $\pm$ \scriptsize0.021 & \textbf{0.766 $\pm$ \scriptsize0.006} \\
    
    stkAAVE & 0.601 $\pm$ \scriptsize0.121 & 0.573 $\pm$ \scriptsize0.084 & 0.517 $\pm$ \scriptsize0.071 & 0.609 $\pm$ \scriptsize0.032 & \underline{0.624 $\pm$ \scriptsize0.017} & \textbf{0.650 $\pm$ \scriptsize0.028} & 0.687 $\pm$ \scriptsize0.101 & 0.616 $\pm$ \scriptsize0.108 & 0.571 $\pm$ \scriptsize0.045 & 0.669 $\pm$ \scriptsize0.066 & 0.710 $\pm$ \scriptsize0.017 & \textbf{0.736 $\pm$ \scriptsize0.022} \\
    
    GLM & 0.448 $\pm$ \scriptsize0.097 & 0.363 $\pm$ \scriptsize0.132 & 0.331 $\pm$ \scriptsize0.083 & \textbf{0.563 $\pm$ \scriptsize0.016} & 0.463 $\pm$ \scriptsize0.053 & 0.502 $\pm$ \scriptsize0.027 & 0.467 $\pm$ \scriptsize0.041 & 0.436 $\pm$ \scriptsize0.052 & 0.437 $\pm$ \scriptsize0.020 & \textbf{0.541 $\pm$ \scriptsize0.010} & 0.480 $\pm$ \scriptsize0.026 & \underline{0.490 $\pm$ \scriptsize0.012} \\
    
    QOM & 0.594 $\pm$ \scriptsize0.043 & 0.613 $\pm$ \scriptsize0.009 & 0.574 $\pm$ \scriptsize0.030 & \underline{0.614 $\pm$ \scriptsize0.005} & 0.614 $\pm$ \scriptsize0.007 & \textbf{0.618 $\pm$ \scriptsize0.004} & 0.587 $\pm$ \scriptsize0.020 & \underline{0.598 $\pm$ \scriptsize0.004} & 0.573 $\pm$ \scriptsize0.018 & 0.596 $\pm$ \scriptsize0.005 & 0.597 $\pm$ \scriptsize0.005 & \textbf{0.599 $\pm$ \scriptsize0.003} \\
    
    WOJAK & 0.516 $\pm$ \scriptsize0.057 & 0.524 $\pm$ \scriptsize0.016 & \underline{0.561 $\pm$ \scriptsize0.026} & 0.489 $\pm$ \scriptsize0.060 & \textbf{0.598 $\pm$ \scriptsize0.075} & 0.534 $\pm$ \scriptsize0.020 & 0.810 $\pm$ \scriptsize0.052 & 0.838 $\pm$ \scriptsize0.014 & 0.834 $\pm$ \scriptsize0.016 & 0.808 $\pm$ \scriptsize0.027 & \textbf{0.862 $\pm$ \scriptsize0.026} & \underline{0.844 $\pm$ \scriptsize0.007} \\
    
    DINO & 0.667 $\pm$ \scriptsize0.138 & 0.695 $\pm$ \scriptsize0.147 & \textbf{0.738 $\pm$ \scriptsize0.047} & 0.617 $\pm$ \scriptsize0.148 & \underline{0.704 $\pm$ \scriptsize0.065} & 0.659 $\pm$ \scriptsize0.039 & 0.719 $\pm$ \scriptsize0.120 & \underline{0.740 $\pm$ \scriptsize0.107} & \textbf{0.746 $\pm$ \scriptsize0.083} & 0.619 $\pm$ \scriptsize0.073 & 0.683 $\pm$ \scriptsize0.046 & 0.643 $\pm$ \scriptsize0.041 \\
    
    Metis & \underline{0.692 $\pm$ \scriptsize0.023} & 0.677 $\pm$ \scriptsize0.030 & 0.609 $\pm$ \scriptsize0.025 & 0.674 $\pm$ \scriptsize0.020 & 0.690 $\pm$ \scriptsize0.016 & \textbf{0.697 $\pm$ \scriptsize0.01}3 & 0.558 $\pm$ \scriptsize0.029 & 0.541 $\pm$ \scriptsize0.083 & 0.485 $\pm$ \scriptsize0.065 & 0.555 $\pm$ \scriptsize0.019 & \textbf{0.586 $\pm$ \scriptsize0.022} & \underline{0.564 $\pm$ \scriptsize0.019} \\
    
    REPv2 & 0.670 $\pm$ \scriptsize0.053 & 0.686 $\pm$ \scriptsize0.043 & 0.706 $\pm$ \scriptsize0.040 & \textbf{0.735 $\pm$ \scriptsize0.017} & 0.707 $\pm$ \scriptsize0.019 & \underline{0.733 $\pm$ \scriptsize0.019} & 0.617 $\pm$ \scriptsize0.080 & 0.633 $\pm$ \scriptsize0.008 & 0.619 $\pm$ \scriptsize0.054 & \textbf{0.720 $\pm$ \scriptsize0.031} & 0.654 $\pm$ \scriptsize0.032 & \underline{0.683 $\pm$ \scriptsize0.022} \\
    
    TRAC & 0.736 $\pm$ \scriptsize0.015 & 0.736 $\pm$ \scriptsize0.014 & 0.710 $\pm$ \scriptsize0.027 & \underline{0.741 $\pm$ \scriptsize0.003} & 0.741 $\pm$ \scriptsize0.007 & \textbf{0.742 $\pm$ \scriptsize0.004} & 0.702 $\pm$ \scriptsize0.031 & 0.709 $\pm$ \scriptsize0.020 & 0.708 $\pm$ \scriptsize0.016 & \textbf{0.720 $\pm$ \scriptsize0.002} & \underline{0.717 $\pm$ \scriptsize0.003} & 0.717 $\pm$ \scriptsize0.005 \\
    
    BEPRO & 0.723 $\pm$ \scriptsize0.053 & 0.720 $\pm$ \scriptsize0.035 & 0.685 $\pm$ \scriptsize0.069 & 0.734 $\pm$ \scriptsize0.016 & \textbf{0.757 $\pm$ \scriptsize0.015} & \underline{0.746 $\pm$ \scriptsize0.015} & 0.730 $\pm$ \scriptsize0.096 & 0.755 $\pm$ \scriptsize0.019 & 0.727 $\pm$ \scriptsize0.063 & 0.764 $\pm$ \scriptsize0.021 & \textbf{0.791 $\pm$ \scriptsize0.007} & \underline{0.776 $\pm$ \scriptsize0.012} \\

\bottomrule
   
    \end{tabular}%
    }
\end{table*}

\begin{table*}[ht]

    \centering
 \caption{\textcolor{black}{\textbf{AUC} scores of multi-network models, single models, and persistence forecasts on train sets across three seeds. The best performance is shown in bold.}}
\label{tab:train_set}
  \resizebox{\linewidth}{!}{%
      \begin{tabular}{l | c | c  | c || l | c | c  | c}
    \toprule 
  
    Token  & Per. Fore. & Single Model & \methodname-64 & Token  & Per. Fore. & Single Model & \methodname-64   \\ \midrule
    LADYS & 0.324 & 0.359 $\pm$ \scriptsize0.127 & \textbf{0.744 $\pm$ \scriptsize0.020} & PUSH & 0.450 & 0.635 $\pm$ \scriptsize0.010 & \textbf{0.657 $\pm$ \scriptsize0.027} \\
    
    Mog & 0.333 & 0.426 $\pm$ \scriptsize0.042 & \textbf{0.708 $\pm$ \scriptsize0.066} & CMT & 0.262 & 0.726 $\pm$ \scriptsize0.095 & \textbf{0.737 $\pm$ \scriptsize0.031} \\
    
    STARL & 0.219 & 0.444 $\pm$ \scriptsize0.005 & \textbf{0.677 $\pm$ \scriptsize0.023} & SLP & 0.415 & 0.500 $\pm$ \scriptsize0.015 & \textbf{0.510 $\pm$ \scriptsize0.022} \\
    
    LQTY & 0.366 & 0.533 $\pm$ \scriptsize0.218 & \textbf{0.729 $\pm$ \scriptsize0.009} & LUSD & 0.372 & 0.722 $\pm$ \scriptsize0.013 & \textbf{0.729 $\pm$ \scriptsize0.08} \\
    
    WOOL & 0.507 & 0.607 $\pm$ \scriptsize0.063 & \textbf{0.776 $\pm$ \scriptsize0.058} & LINA & 0.428 & 0.749 $\pm$ \scriptsize0.052 & \textbf{0.755 $\pm$ \scriptsize0.005} \\
    
    BITCOIN & 0.382 & 0.586 $\pm$ \scriptsize0.040 & \textbf{0.744 $\pm$ \scriptsize0.042} & AIOZ & 0.39 & 0.745 $\pm$ \scriptsize0.013 & \textbf{0.745 $\pm$ \scriptsize0.012} \\
    
    aDAI & 0.434 & 0.528 $\pm$ \scriptsize0.033 & \textbf{0.657 $\pm$ \scriptsize0.045} & sILV2 & 0.581 & \textbf{0.886 $\pm$ \scriptsize0.001} & 0.602 $\pm$ \scriptsize0.023 \\

    RSR & 0.542 & 0.530 $\pm$ \scriptsize0.016 & \textbf{0.649 $\pm$ \scriptsize0.028} & INU & 0.292 & \textbf{1.000 $\pm$ \scriptsize0.000} & 0.793 $\pm$ \scriptsize0.167 \\

    DRGN & 0.385 & 0.561 $\pm$ \scriptsize0.095 & \textbf{0.660 $\pm$ \scriptsize0.005} & SPONGE & 0.167 & \textbf{0.650 $\pm$ \scriptsize0.088} & 0.458 $\pm$ \scriptsize0.027 \\

    FEG & 0.442 & 0.500 $\pm$ \scriptsize0.025 & \textbf{0.593 $\pm$ \scriptsize0.011} & bendWETH & 0.49 & \textbf{0.651 $\pm$ \scriptsize0.047} & 0.522 $\pm$ \scriptsize0.019 \\

    SWAP & 0.468 & 0.633 $\pm$ \scriptsize0.013 & \textbf{0.724 $\pm$ \scriptsize0.005} & TNT & 0.469 & \textbf{0.796 $\pm$ \scriptsize0.006} & 0.700 $\pm$ \scriptsize0.027 \\

    POLS & 0.393 & 0.736 $\pm$ \scriptsize0.022 & \textbf{0.818 $\pm$ \scriptsize0.007} & CRU & 0.431 & \textbf{0.793 $\pm$ \scriptsize0.053} & 0.698 $\pm$ \scriptsize0.007 \\

    ORN & 0.333 & 0.669 $\pm$ \scriptsize0.08 & \textbf{0.747 $\pm$ \scriptsize0.009} & AUDIO & 0.441 & \textbf{0.789 $\pm$ \scriptsize0.023} & 0.694 $\pm$ \scriptsize0.006 \\

    HOP & 0.415 & 0.664 $\pm$ \scriptsize0.170 & \textbf{0.738 $\pm$ \scriptsize0.021} & ALBT & 0.317 & \textbf{0.734 $\pm$ \scriptsize0.01} & 0.642 $\pm$ \scriptsize0.009 \\

    RFD & 0.277 & 0.721 $\pm$ \scriptsize0.021 & \textbf{0.794 $\pm$ \scriptsize0.021} & MAHA & 0.284 & \textbf{0.902 $\pm$ \scriptsize0.024} & 0.815 $\pm$ \scriptsize0.031 \\

    aUSDC & 0.513 & 0.685 $\pm$ \scriptsize0.009 & \textbf{0.757 $\pm$ \scriptsize0.016} & YFII & 0.315 & \textbf{0.775 $\pm$ \scriptsize0.039} & 0.691 $\pm$ \scriptsize0.024 \\

    steCRV & 0.360 & 0.591 $\pm$ \scriptsize0.035 & \textbf{0.660 $\pm$ \scriptsize0.006} & BTRFLY & 0.127 & \textbf{0.846 $\pm$ \scriptsize0.029} & 0.765 $\pm$ \scriptsize0.039 \\

    ANT & 0.469 & 0.629 $\pm$ \scriptsize0.039 & \textbf{0.695 $\pm$ \scriptsize0.010} & RLB & 0.273 & \textbf{0.963 $\pm$ \scriptsize0.032} & 0.887 $\pm$ \scriptsize0.042 \\

    SUPER & 0.432 & 0.729 $\pm$ \scriptsize0.010 & \textbf{0.794 $\pm$ \scriptsize0.000} & crvUSD & 0.291 & \textbf{0.417 $\pm$ \scriptsize0.126} & 0.344 $\pm$ \scriptsize0.025 \\

    ShibDoge & 0.514 & 0.768 $\pm$ \scriptsize0.034 & \textbf{0.832 $\pm$ \scriptsize0.006} & KP3R & 0.528 & \textbf{0.839 $\pm$ \scriptsize0.033} & 0.775 $\pm$ \scriptsize0.016 \\

    0x0 & 0.383 & 0.555 $\pm$ \scriptsize0.158 & \textbf{0.616 $\pm$ \scriptsize0.018} & CELR & 0.495 & \textbf{0.782 $\pm$ \scriptsize0.011} & 0.722 $\pm$ \scriptsize0.032 \\

    NOIA & 0.359 & 0.567 $\pm$ \scriptsize0.042 & \textbf{0.626 $\pm$ \scriptsize0.003} & INJ & 0.444 & \textbf{0.741 $\pm$ \scriptsize0.093} & 0.686 $\pm$ \scriptsize0.034 \\

    REP & 0.360 & 0.711 $\pm$ \scriptsize0.116 & \textbf{0.769 $\pm$ \scriptsize0.019} & QSP & 0.431 & \textbf{0.695 $\pm$ \scriptsize0.006} & 0.647 $\pm$ \scriptsize0.006 \\

    DODO & 0.346 & 0.717 $\pm$ \scriptsize0.022 & \textbf{0.774 $\pm$ \scriptsize0.013} & RGT & 0.396 & \textbf{0.864 $\pm$ \scriptsize0.056} & 0.826 $\pm$ \scriptsize0.007 \\

    PSYOP & 0.403 & 0.848 $\pm$ \scriptsize0.034 & \textbf{0.904 $\pm$ \scriptsize0.018} & ARC & 0.532 & \textbf{0.760 $\pm$ \scriptsize0.004} & 0.736 $\pm$ \scriptsize0.020 \\

    TURBO & 0.789 & 0.840 $\pm$ \scriptsize0.012 & \textbf{0.893 $\pm$ \scriptsize0.018} & BOB & 0.105 & \textbf{0.717 $\pm$ \scriptsize0.026} & 0.696 $\pm$ \scriptsize0.053 \\

    Yf-DAI & 0.434 & 0.725 $\pm$ \scriptsize0.006 & \textbf{0.777 $\pm$ \scriptsize0.002} & GHST & 0.344 & \textbf{0.734 $\pm$ \scriptsize0.050} & 0.726 $\pm$ \scriptsize0.015 \\

    cDAI & 0.519 & 0.566 $\pm$ \scriptsize0.229 & \textbf{0.606 $\pm$ \scriptsize0.014} & PICKLE & 0.321 & \textbf{0.838 $\pm$ \scriptsize0.031} & 0.831 $\pm$ \scriptsize0.011 \\

    POOH & 0.250 & 0.755 $\pm$ \scriptsize0.266 & \textbf{0.786 $\pm$ \scriptsize0.019} & IOTX & 0.366 & \textbf{0.706 $\pm$ \scriptsize0.039} & 0.699 $\pm$ \scriptsize0.019 \\

    PRE & 0.481 & 0.673 $\pm$ \scriptsize0.024 & \textbf{0.702 $\pm$ \scriptsize0.010} & RARI & 0.440 & \textbf{0.750 $\pm$ \scriptsize0.019} & 0.744 $\pm$ \scriptsize0.011 \\

    TVK & 0.376 & 0.711 $\pm$ \scriptsize0.014 & \textbf{0.738 $\pm$ \scriptsize0.005} & DPI & 0.291 & \textbf{0.751 $\pm$ \scriptsize0.014} & 0.746 $\pm$ \scriptsize0.012 \\

    OHM & 0.652 & 0.617 $\pm$ \scriptsize0.004 & \textbf{0.643 $\pm$ \scriptsize0.011} & MIM & 0.372 & \textbf{0.698 $\pm$ \scriptsize0.010} & 0.697 $\pm$ \scriptsize0.015 \\
    
\bottomrule
   
    \end{tabular}%
    }
\end{table*}

\include{tables/table3_rerun}

\end{document}